\documentclass[12pt,letterpaper]{article}
\usepackage{float}
\newcommand {\be}{\begin{equation}}
\newcommand {\ee}{\end{equation}}
\newcommand {\bea}{\begin{eqnarray}}
\newcommand {\eea}{\end{eqnarray}}

\usepackage[dvips]{graphicx}
\graphicspath{{../eps/}}
\DeclareGraphicsExtensions{.eps}
\usepackage{epsfig,color}
\usepackage{booktabs}
\usepackage{epstopdf}
\usepackage{amssymb}
\usepackage{mathtools,cite}
\usepackage{xspace} 
\usepackage{algorithmic}
\usepackage[section]{placeins}
\usepackage{array}

\usepackage{mdwmath}
\usepackage{mdwtab}
\usepackage{bm}
\usepackage{mathrsfs}
\usepackage{epsfig}
\usepackage{subfigure}
\usepackage{algorithmic}
\usepackage{amsmath}
\usepackage{mathrsfs}
\usepackage{cleveref}  
\usepackage[ruled, lined, linesnumbered, commentsnumbered, longend]{algorithm2e}
\usepackage{xcolor}

\newcommand{\resnet}{{ResNet-18}\xspace}
\newcommand{\densenet}{{DenseNet-121}\xspace}
\begin{document}

\begin{centering}
{\large \bf
Activity-weight duality in feed forward neural networks reveals two co-determinants for generalization}\\

\vspace{0.5cm}
Yu Feng$^{1,2}$, Wei Zhang$^{1}$, and Yuhai Tu$^{1}$\\
$^{1}$IBM T. J. Watson Research Center, NY 10598\\
$^{2}$Department of Physics, Duke University, Durham, NC 27710\\

\end{centering}
\vspace{1cm}

\begin{center}
{\bf Abstract}
\end{center}
Generalization is a fundamental problem in machine learning especially for overparameterized neural networks where there are many weight solutions that fit the training data equally well. Here, we report the discovery of duality relations between changes in activities in a densely connected layer of neurons and the changes in their weights connecting to the next layer, which allows us to decompose the overall generalization loss into contributions from different directions in weight space. We find that the loss from each direction is the product of two geometric factors (determinants): sharpness of the loss landscape and the standard deviation of the dual weights, which scales as an activity-weighted norm of the solution. By using the generalization loss decomposition, we uncover how different regularization schemes affect generalization by controlling one or both factors. We applied our analysis framework to evaluate different algorithms for realistic large neural network models in the multi-learner setting. We found that the decentralized algorithms have better generalization performance as they introduce additional landscape-dependent noise that leads to flatter solutions without changing their norms.
\newpage

\section{Introduction}

Generalization is one of the most important problems in machine learning. This problem becomes more pressing given the overwhelming number of parameters (weights) used in feed forward deep learning neural networks (DLNN) ~\cite{LeCun2015Deep,goodfellow2016deep}, which have enjoyed a long string of tremendous successes in achieving human level performance in image recognition~\cite{he2016deep}, machine translation~\cite{wu2016google}, games~\cite{alpha-go}, and even solving longstanding grand challenge scientific problems such as protein folding~\cite{AlphaFold}. Specifically, given the large number of parameters in DLNN, there are many different solutions that can fit the training data equally well. Ultimately, the ``quality" of a solution is determined by its generalizability, i.e., how well the solution fits a test data set that the neural network has not seen before~\cite{zhang2016understanding}.  Thus, the key question is what properties of a solution makes it more generalizable.

There has been much recent work on generalization in DLNN based on various theoretically and empirically motivated complexity measures (VC-dimension, norm of parameters, sharpness, path norms, etc.) of the solution, see Jiang et al~\cite{jiang2019fantastic} for a recent comprehensive review on the subject and the references therein. However, despite empirical evidence for a strong correlation between sharpness-based measures and generalization~\cite{KeskarMNST17}, the other (theoretically motivated) measures such as the norm-based measures do not serve as robust indicators for generalization by themselves~\cite{jiang2019fantastic}. Even in the case of the more promising sharpness-based measures, we do not understand exactly why and how they are effective in predicting generalization. Furthermore, there are recent work challenging the validity of using loss landscape sharpness alone for determining generalization based on a general scaling invariance in DLNN~\cite{dinh2017sharp}. Indeed, a comprehensive understanding of generalization in DLNN still remains elusive. 

Here, we tackle the generalization problem by using a different approach. The general motivation is that if there exists an equivalence (duality) between the change of the input between a training data ($\boldsymbol{x}$) and a testing data ($\boldsymbol{x'}$) and a corresponding change of the weights from the solution ($\boldsymbol{w}$) to a new weight ($\boldsymbol{w'}$), we can then use this duality to map a distribution in the input space to a distribution in the weight space, where we can evaluate the generalization loss. Remarkably, we find an infinite family of such exact equivalence (duality) relations between changes in activities and weights in any densely connected layer (number of independent weights per neuron is larger than $1$) in feed forward neural networks. By using the ``minimal" duality relation with the smallest weight change, we decompose the generalization loss to contributions from different eigen-directions of the Hessian matrix of the loss function at the solution in the weight space. The form of these contributions reveals two co-acting factors (determinants) for generalization -- one is governed by the sharpness of the loss landscape and the other corresponds to the norm of the solution weighted by the covariance of the relative differences between the training and testing data set. 

The primary goal of our study is to understand the underlying reason(s) for the generalizability of different solutions for the same task. Indeed, the analytical expression for the generalization gap obtained by using the activity-weight duality allows us to compare the generalization performances of different solutions reached by different algorithms; the same algorithm with different hyperparameters and/or different regularization schemes. In particular, the exact decomposition of the generalization gap in different weight directions allows us to explain how these implicit and explicit regularization schemes as well as training data size and mislabeled data affect generalization by varying one or both of the two generalization determinants of the solution (sharpness and size).  Finally, to demonstrate the utility of our theoretical results, we apply our analysis framework to evaluate the generalization performance of different SGD-based algorithms in realistic large neural network models with multiple learners (processors). We find that the decentralized parallel SGD algorithm has better generalization than the synchronized SGD due to its additional landscape-dependent noise, which leads to flatter solutions.    

\section{Results}

\subsection{The activity-weight duality in feed-forward neural networks}

In a neural network, the loss function $l_k$ for an individual sample-$k$ depends on both the input $\boldsymbol{x}_k$, the output $\boldsymbol{y}_k$,  and parameters of the model given by all the weights ($\boldsymbol{W}\equiv (\boldsymbol{w}^{[1]}, \boldsymbol{w}^{[2]}, ..., \boldsymbol{w}^{[s]}, ..., \boldsymbol{w}^{[S]}) $) of the network where $\boldsymbol{w}^{[s]}$ are the weights between layer-$s$ and layer-$(s-1)$ and $S$ is the total number of layers in the network excluding the input layer, which can be considered as layer-$0$ ($s=0$). The solutions for the weights are learned by minimizing the average training loss $L(\boldsymbol{W})=\langle l_k\rangle_{k\in\mathcal{D}_{tr}}$ where the average is taken over the training sample-$k$ that belongs to the training set  $\mathcal{D}_{tr}$. Given the overparametrized nature of deep nets, there are many solutions with very small or near zero training loss. However, the performance of the neural network model is determined by how well the model fits the previously unseen test dataset $\mathcal{D}_{te}$ as characterized by the test loss $L'(\boldsymbol{W})=\langle l_k\rangle_{k\in\mathcal{D}_{te}}$. Indeed, generalizability of a given solution $\boldsymbol{W}$ can be measured by the generalization gap $\Delta L(\boldsymbol{W})\equiv L'(\boldsymbol{W})-L(\boldsymbol{W})$. The smaller the gap the more generalizable the solution is. 

\begin{figure}[htbp]
\centering
\includegraphics[clip, trim=5cm 7.5cm 4.5cm 7cm, width=1\linewidth]{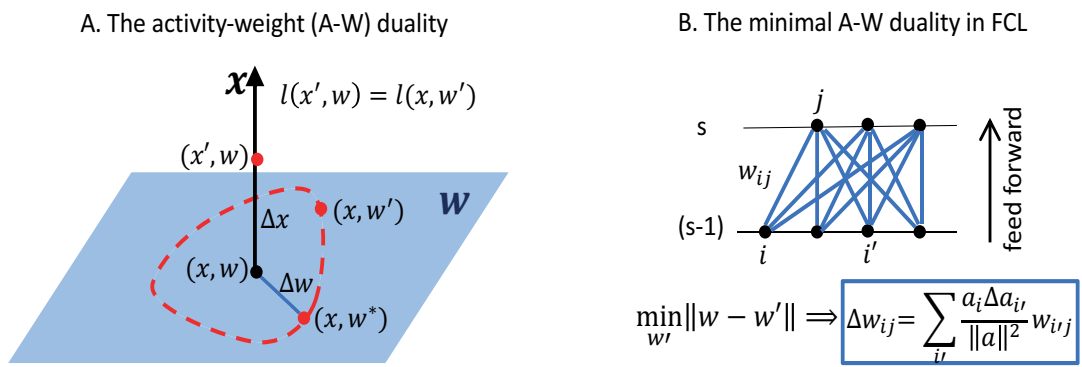}
\caption{The activity-weight duality. (A) Illustration of the duality relation. The activity and weight are represented by the vertical direction and the horizontal plane (shaded blue), respectively. The loss for a new activity $\boldsymbol{x'}$ at the original weight $\boldsymbol{w}$ (red dot on the vertical axis) is the same as the loss at the original activity ($\boldsymbol{x}$) at a new weight $\boldsymbol{w'}$, which can be any point on the dotted red line in the weight($\boldsymbol{w}$)-plane. (B) Between two fully connected layers ($(s-1)$ and $s$) in a feed-forward network, changes of weights in layer-$s$ ($\boldsymbol{\Delta w}$) for the duality relation with the minimum $||\boldsymbol{\Delta w}||_2$ are given explicitly by the original weights ($\boldsymbol{w}$), the original activity ($\boldsymbol{a}$) and the changes in activity ($\boldsymbol{\Delta a}$) in layer-$(s-1)$).}
\label{fig:lr1} 
\end{figure}

{\bf Motivation for the activity-weight duality.} The generalization gaps for different solutions can be computed directly to decide which of them is more generalizable. However, such direct comparison does not reveal the underlying reason(s) why one solution is more generalizable than the other. To search for the key underlying factors that contribute to the generalization gap, we look for a possible data-parameter duality relation: 
\begin{equation}
l(\boldsymbol{x'},\boldsymbol{W})=l(\boldsymbol{x},\boldsymbol{W'}),
\label{dual}
\end{equation} 
which means that in terms of the loss function a change from a training data $\boldsymbol{x}$ to a test data $\boldsymbol{x'}$ with fixed parameters $\boldsymbol{W}$ is equivalent to changing the parameters from $\boldsymbol{W}$ to $\boldsymbol{W'}$ without changing the data (see Fig.~\ref{fig:lr1}A for an illustration). Here, we call $\boldsymbol{W'}$ the dual weight corresponding to the data-pair ($\boldsymbol{x}'$,$\boldsymbol{x}$). 

Why is such a duality relation helpful for understanding generalization? The reason is that an activity-weight duality allows us to determine the generalization gap of a given solution by just evaluating the training loss landscape $L(\boldsymbol{W})$ in the neighborhood of that solution in the weight space. Specifically, we can determine the generalization gap by computing the difference in training losses at different weights -- one at the solution itself (for the training data) and the other at the dual weight (for the test data). As we will show below, the dual weight $\boldsymbol{W}'$ is data-dependent and all the dual weights form a distribution of weights centered around the solution $\boldsymbol{W}$ . This immediately suggests that the generalization gap depends on the span of the dual weight distribution as well as the shape of the training loss landscape around the solution $\boldsymbol{W}$, which can be characterized by the flatness of the training loss landscape near the solution which is at a minimum of the training loss landscape.         

{\bf The minimal activity-weight duality relation in a fully connected layer.} Given the much larger number of weights than the dimension of data, in principle there should exist many solutions to the dual weight ($\boldsymbol{W}'$) for a given pair of train-test data ($\boldsymbol{x},\boldsymbol{x}'$). However, $\boldsymbol{W}'$ may be hard to find explicitly and they may not have the right structure for understanding generalization. Below, we construct an explicit solution of the dual weights $\boldsymbol{W}'$ where only the weights $\boldsymbol{w}^{[(s)]}$ between two fully connected layers, i.e., layer-$(s-1)$ and layer-$s$, are changed while the rest of the weights remain the same as the original weights $\boldsymbol{W}$, i.e., $\boldsymbol{w}'^{[x]}=\boldsymbol{w}^{[x]}$ for all other layers $x(\ne s)$. 

For the two fully connected layers, the activity of a neuron $j$ in the layer-$s$ is determined by the pre-activation factor $z^{[s-1]}_{j}=\sum_j w^{[s]}_{ij}a^{[s-1]}_i$ where $a^{[s-1]}_i$ is the activity of neuron-$i$ in the preceding layer-$(s-1)$. Note that the input $\boldsymbol{x}$ can be considered as the neuron activities in the input layer (layer-$0$): $\boldsymbol{x}=\boldsymbol{a}^{[0]}$. 
If there is no skip connection from layers below layer-$s$ to layers above layer-$s$, the loss function in feed-forward neural networks only depends on the activities of neurons in layer-$s$: $a^{[s]}_j=\phi(z^{[s-1]}_j)$ with $\phi(\cdot)$ the activation function, and the weights above layer-$s$. Therefore, the duality relation Eq.~\ref{dual} can be satisfied if we change only the weights $\boldsymbol{w}^{[s]}$ between layer-$(s-1)$ and layer-$s$ to keep the pre-activation $\boldsymbol{z}^{[s-1]}$ unchanged so that  $\boldsymbol{a}^{[s]}$ remain unchanged. The dual weights $\boldsymbol{w}'$ can thus be determined by solving the linear equations: 
\begin{equation}
\sum_{i=1}^{H_1}w'^{[s]}_{ij} a_i^{[s-1)]} =\sum_{i=1}^{H_1}w_{ij}^{[s]} a'^{[s-1]}_i,\;\;\; j=1,2,...,H_2
\label{constrain}
\end{equation}
where $H_1$ and $H_2$ are the number of neurons in layer-$(s-1)$ and layer-$s$, respectively.

For simplicity, we will drop the superscript for layer identification in the rest of this paper (unless otherwise stated). Specifically,  $\boldsymbol{w}$ and $\boldsymbol{w}'$ refer to the weights and the dual weights between two fully connected layers ($(s-1)$ and $s$) as illustrated in Fig.~\ref{fig:lr1}B; 
$\boldsymbol{a}$ and $\boldsymbol{a}'$ correspond to the neuron activities in layer-$(s-1)$ for a training data ($\boldsymbol{x}$) and a testing data ($\boldsymbol{x}'$), respectively. Both $\boldsymbol{a}$ and $\boldsymbol{a}'$ depend on all the weights preceding layer-$(s-1)$: $\boldsymbol{a}=\boldsymbol{\mathcal{G}}(\boldsymbol{x},\boldsymbol{w}^{[1]},\boldsymbol{w}^{[2]}, ...,\boldsymbol{w}^{[s-1]})$, $\boldsymbol{a}'=\boldsymbol{\mathcal{G}}(\boldsymbol{x}',\boldsymbol{w}^{[1]},\boldsymbol{w}^{[2]}, ...,\boldsymbol{w}^{[s-1]})$ where $\boldsymbol{\mathcal{G}}$ represents the network model up to layer-$(s-1)$. In practice, they can be easily computed by propagating the network to layer-$(s-1)$ with the original weights and different inputs ($\boldsymbol{x}$ and $\boldsymbol{x}'$).

In a fully connected layer (FCL) considered here, the total number of weights ($M=H_1H_2$) is larger than the number of neurons ($H_2$); therefore there are an infinite number of solutions for $\boldsymbol{w'}$ that satisfies the duality condition (Eq.~\ref{constrain}). Here, we focus on the ``minimal" duality solution $\boldsymbol{w^*}$ that is the closest to the original weights $\boldsymbol{w}$. The minimal duality solution $\boldsymbol{w^*}$ can be obtained by minimizing $||\boldsymbol{w'}-\boldsymbol{w}||^2$ under the constraints given by Eq.~\ref{constrain}. 
This problem can be solved exactly by introducing a Lagrange multiplier $\lambda_j$ for each constraint $j$ and minimizing:
\begin{equation}
\label{Lagrange}
S(w,\lambda) =\sum_{ij}\Delta w_{ij}^2 +\sum_{j}\lambda_j \sum_i ( \Delta w_{ij} a_i -w_{ij} \Delta a_i),
\end{equation}
where $\Delta w_{ij}=w^*_{ij}-w_{ij}$ and $\Delta a_{i}=a'_{i}-a_{i}$. Optimizing Eq.~\ref{Lagrange} leads to the solution:
\begin{equation}
\Delta w_{ij}=-\lambda_j x_i/2,
\end{equation}
and the Lagrange multiplier can be determined by the constraints:
\begin{equation}
\lambda _j =-\frac{2 \sum_{i} w_{ij} \Delta a_i}{\sum_i a_i^2},
\end{equation}
which finally leads to an exact expression of the minimal duality weights $w^*_{ij}= w_{ij}+\Delta w_{ij}$ with the dual weight shift $\Delta w_{ij}$ given as a linear function of the original weights: 
\begin{equation}
\Delta w_{ij}=\sum_{i'=1}^{H_1} b_{i'i} w_{i'j}=\frac{ a_i}{||\boldsymbol{a}||^2}\sum_{i'=1}^{H_1} \Delta a_{i'}w_{i'j},
\label{opt}
\end{equation}
where the linear coefficients $b_{i'i}= \frac{ \Delta a_{i'} a_i}{||\boldsymbol{a}||^2}$ are determined by the overall neuron activity strength ($||\boldsymbol{a}||^2$) and the correlation between $\boldsymbol{a}$ and $ \boldsymbol{\Delta a}(\equiv \boldsymbol{a'}-\boldsymbol{a})$ at different neurons (see Fig.~\ref{fig:lr1}B for an illustration).

\subsection{Decomposition of the generalization gap in weight space}

To use the activity-weight duality for computing the generalization gap, we form the test-train sample pairs.  For each test sample $\boldsymbol{x}'_k$ with $k\in\mathcal{D}_{te}$, we form a test-train pair ($\boldsymbol{x}'_{k}$, $\boldsymbol{x}_k$) where $\boldsymbol{x}_{k}$ is the closest training sample with the same label. For simplicity, we define closeness by using the $L_2$ (Euclidean) distance in this paper. We call $\boldsymbol{x}_{k}$ the training partner sample for $\boldsymbol{x}'_{k}$. The set of training partner samples for all the test samples is called the training partner set $\mathcal{D}_p$. 

The generalization gap between a test-train pair-$k$ is defined as the difference in their losses: $\Delta l_k \equiv l(\boldsymbol{x'}_k,\boldsymbol{w})- l(\boldsymbol{x}_{k},\boldsymbol{w})$. By averaging $\Delta l_k$ over all the test samples $k\in\mathcal{D}_{te}$ and their training partner samples, we define an average ``pair-wise" generalization gap:
\begin{equation}
\Delta \tilde{L} \equiv \langle \Delta l_k\rangle_{k\in\mathcal{D}_{te}}=\langle l_k\rangle_{k\in\mathcal{D}_{te}}-\langle l_{k_p}\rangle_{k\in\mathcal{D}_{p}}=L'-\tilde{L},
\label{DL}
\end{equation}
where $L'$ is the average test loss over the test dataset and $\tilde{L}$ is the average loss over the training samples in the training partner dataset $\mathcal{D}_p$. Since $\mathcal{D}_p$ is a subset of the entire training set $\mathcal{D}_{tr}$, $\tilde{L}$ is not exactly the same as the average training loss $L$. However, given the large sample sizes in $\mathcal{D}_p$, $\tilde{L}$ serves as a good approximation for the average training loss: $\tilde{L}\approx L$. Furthermore, the test loss is typically much larger than the training loss: $L'\gg L\approx\tilde{L}$, therefore, the true generalization gap can be accurately approximated by the pair-wise generalization gap: $\Delta L\approx \Delta \tilde{L}$. In fact, this approximation remains valid as long as the pair-wise sample size is large enough, i.e., it does not need to cover the whole test dataset, see Fig.~\ref{fig:pairloss} in the Supplementary Information (SI) for details. 

In the rest of the paper, we will approximate the generalization gap by using $\Delta\tilde{L}$, which can be decomposed in the weight space. By using the A-W duality relation,
we can rewrite the individual generalization gap as the difference in loss for the same training data ($\boldsymbol{x}_{k}$) but at two different weights: the solution $\boldsymbol{w}$, which only depends on the training data and the dual weights $\boldsymbol{w^*}_k=\boldsymbol{w}+\boldsymbol{\Delta w}_k$, which depends on both test and training data as shown by Eq.~\ref{opt}. 
From the weight shift vector $\boldsymbol{\Delta w}_k$, an ``effective gradient" $\boldsymbol{g}_k$ (vector) can be defined as:
\begin{equation}
\boldsymbol{g}_k\equiv \frac{\Delta l_k}{||\boldsymbol{\Delta w}_k||^2}\boldsymbol{\Delta w}_k.
\label{g_k}
\end{equation}  
By using $\boldsymbol{g}_k$, we can write the generalization gap for sample-$k$ as:
\begin{equation}
\Delta l_k \equiv l(\boldsymbol{x'}_k,\boldsymbol{w})-l(\boldsymbol{x}_k,\boldsymbol{w})=l(\boldsymbol{x}_k,\boldsymbol{w^*}_k)-l(\boldsymbol{x}_k,\boldsymbol{w})
=\boldsymbol{g}_k\cdot \boldsymbol{\Delta w}_k.
\end{equation}

Next, we express the two vectors $\boldsymbol{g}_k$ and $\boldsymbol{\Delta w}_k$ in an orthogonal basis $\{\boldsymbol{e}_n\}$ with $\boldsymbol{e}_n$ the unit vector in direction $n(=1,2,...,M)$: $\boldsymbol{g}_k=\sum_{n=1}^M g_{n,k}\boldsymbol{e}_n$ and $\boldsymbol{\Delta w}_k=\sum_{n=1}^M \Delta w_{n,k}\boldsymbol{e}_n$ with the components given by $g_{n,k}=\boldsymbol{g}_k\cdot \boldsymbol{e}_n$ and $\Delta w_{n,k}=\boldsymbol{\Delta w}_k\cdot \boldsymbol{e}_n$.   
Here, we can use the eigen-directions of the Hessian matrix ($\boldsymbol{\mathcal{H}}=\nabla\nabla L|_w$) of the overall training loss function ($L$) evaluated at a solution $\boldsymbol{w}$ (another choice is the Fisher information matrix) as the basis. By using this basis, we can decompose the generalization gap in the weight space:
\begin{equation}
\Delta L\approx \Delta \tilde{L} \equiv \langle \Delta l_k\rangle_{k} =\sum_{n=1}^M  \langle g_{n,k} \Delta w_{n,k}\rangle_k= \sum_{n=1}^M [c_n \sigma_{g,n} \sigma_{w,n} + \mu_{g,n}\mu_{w,n}],
\label{DL}
\end{equation}
where $\mu_{g,n} \equiv \langle g_{n,k}\rangle_k$ and  $\mu_{w,n} \equiv \langle \Delta w_{n,k}\rangle_k$ are the average components; $\sigma^2_{g,n}\equiv  \langle g_{n,k}^2\rangle_k -\mu_{g,n}^2 $ and $\sigma^2_{w,n}\equiv  \langle \Delta w_{n,k}^2\rangle_k -\mu_{w,n}^2$ are the corresponding variances; and $c_n$ is the correlation coefficient between $g_{n,k}$ and $\Delta w_{n,k}$: $
c_n\equiv \langle (g_{n,k}-\mu_{g,n})(\Delta w_{n,k}-\mu_{w,n})\rangle_k/(\sigma_{g,n} \sigma_{w,n})$.

Since the training and testing samples are from the same distribution, there is an approximate reflection symmetry ($\Delta\boldsymbol{a}\rightarrow -\Delta\boldsymbol{a}$) in the distribution for the activity difference $\Delta\boldsymbol{a}$ for large test sample size $N_{te}$. From Eq.~\ref{opt}, $\Delta \boldsymbol{w}$ depends linearly on the relative activity difference $\delta \boldsymbol{a}\equiv \frac{\Delta\boldsymbol{a}}{||\boldsymbol{a}||}$, whose overall magnitude can be characterized by a small parameter $\epsilon\equiv \langle ||\delta \boldsymbol{a}_k||\rangle_k$ averaged over all samples $k\in\mathcal{D}_{te}$. To the leading order of $\epsilon$, the distribution for $\delta\boldsymbol{a}$ also has the reflection symmetry ($\delta\boldsymbol{a}\rightarrow -\delta\boldsymbol{a}$). As a result, we have $\mu_{w,n} \approx N_{te}^{-1/2}\mathcal{O}(\epsilon)+\mathcal{O}(\epsilon^2)$ where the coefficient for the leading order term scales as $N_{te}^{-1/2}$ due to the reflection symmetry. On the other hand, the standard deviation ($\sigma_{w,n}$) reaches a constant of order $\mathcal{O}(\epsilon)$ as $N_{te}\rightarrow \infty$. Therefore, given that $N_{te}\gg 1$ and $\epsilon \ll 1$, we have $|\mu_{w,n}|/\sigma_{w,n}\sim \mathcal{O}(\epsilon)$, which means $|\mu_{w,n}|\ll \sigma_{w,n}$. These results are verified by direct numerical simulations shown in Fig.~\ref{fig:mu_g&w} in SI. Similar results have been observed that show $|\mu_{g,n}|\ll\sigma_{g,n}$, which are also shown in Fig.~\ref{fig:mu_g&w} in SI.

By neglecting the $\mu_{g,n}\mu_{w,n}$ term in Eq.~\ref{DL}, the overall generalization gap can be decomposed into contributions ($\Delta L_n$) from different directions in the weight-space:
\begin{equation}
\Delta L(\boldsymbol{w}) \approx  \sum_{n=1}^M\Delta L_n (\boldsymbol{w}) = \sum_{n=1}^M c_n\sigma_{w,n}\sigma_{g,n},
\label{GG}
\end{equation}
where each contribution $\Delta L_n=c_n\sigma_{w,n}\sigma_{g,n}$ is proportional to the standard deviation of gradients ($\sigma_{g,n}$) and the standard deviation of the dual weights ($\sigma_{w,n}$) among individual samples with a positive definite coefficient $0\le c_n\le 1$, which does not show strong dependence on $n$ (see Fig.~\ref{fig:c_n} in SI). 
Note that the expression for the generalization gap given in Eq.~\ref{GG} is quantitatively accurate and $\Delta L$ is highly correlated with the test error $\epsilon_{te}$, both of which are verified numerically as shown in Fig.~\ref{fig:test} in the SI. 

\subsection{The two co-acting determinants for generalization and their interpretations}

In an over-parameterized deep network, there are many solutions for a given training dataset. However, their generalization performances measured by their generalization gaps can be different. The decomposition given in Eq.~\ref{GG} allows us to understand where the difference in $\Delta L$ comes from and how it depends on the two determining factors (determinants) $\sigma_{g,n}$ and $\sigma_{w,n}$. Before we present the results on generalization, we first explain the geometrical interpretations of $\sigma_{g,n}$ and $\sigma_{w,n}$ and how they relate to the sharpness and size of the solution.  
\begin{figure}[htbp]
\centering
\includegraphics[width=0.8\linewidth]{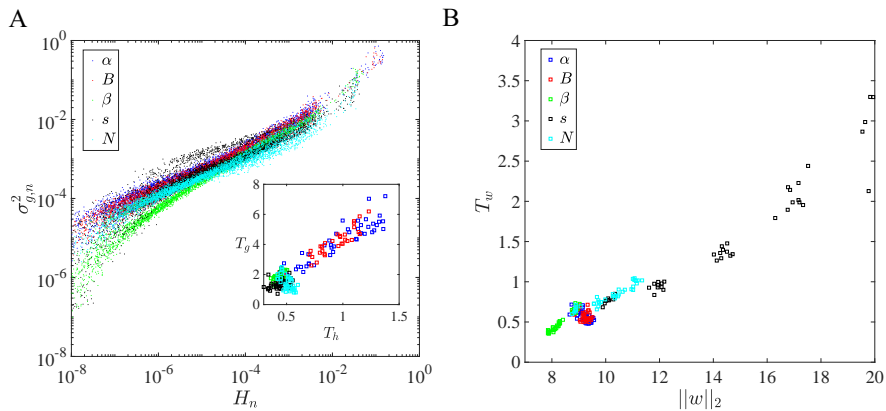}
\caption{Characteristics of $\sigma_{g,n}$ and $\sigma_{w,n}$. (A) $\sigma^2_{g,n}$ versus Hessian eigenvalue $H_n$ for all hyperparameters (learning rate $\alpha$ and batch size $B$ in SGD; decay rate $\beta$ and initial weight scale $s$) and sample size ($N$) studied in this paper. The inset shows that the total gradient variance $T_g\equiv \sum_{n=1}^M \sigma_{g,n}^2$ is highly correlated with the trace of the Hessian ($T_h\equiv \sum_{n=1}^M H_n$). (B) The total dual weight variance $T_w\equiv \sum_{n=1}^M \sigma^2_{w,n}$ versus the $L_2$-norm of the weight $||\boldsymbol{w}||_2$. Both the $L_2$ weight norm and $S_w$ remain roughly constant when $\alpha$ and $B$ are varied. When $\beta$ and $s$ are varied, $||\boldsymbol{w}||_2$ changes significantly, and $T_w$ increases with $||\boldsymbol{w}||_2$.}
\label{fig:sigmas}
\end{figure}

{\bf Geometrical interpretation of $\sigma_{g,n}$.} From its definition  given in Eq.~\ref{g_k}, $g_{k,n}$ is a difference quotient of the loss function for sample-$k$ along direction-$n$ with a finite difference $\Delta w_{k,n}$, therefore its variance $\sigma_{g,n}^2$ over all the samples should depend on the sharpness of the overall landscape along direction-$n$. This dependence can be seen intuitively by taking the limit $\Delta w \rightarrow 0$ when $g_{k,n}$ becomes the gradient $g_{k,n}\rightarrow g^{(0)}_{n,k}=\frac{\partial l_k}{\partial w_n}|_w $.  In this limit, $\sigma^2_{g,n}\approx \langle (\frac{\partial l_k}{\partial w_n})^2\rangle_k$, which is just the diagonal element of the expected (empirical) Fisher information matrix $\boldsymbol{\mathcal{F}}\equiv \langle \nabla l_k\nabla l_k\rangle_k$. 
Under certain approximations $\boldsymbol{\mathcal{F}}$ is proportional to the Hessian matrix ($\boldsymbol{\mathcal{H}}\equiv \nabla\nabla L$) at the solution~\cite{zhu2019anisotropic, martens2014new} (see section 1 in SI for details of the derivation), which leads to $\langle(g^{(0)}_{n,k})^2\rangle_k\propto H_n$ with $H_n$ the $n$-th eigenvalue of $\boldsymbol{\mathcal{H}}$. 

Since $\Delta w_{k,n}$ is finite, the exact quantitative dependence of $\sigma_{g,n}^2$ on $H_n$ can be different from that obtained at the $\Delta \boldsymbol{w}\rightarrow 0$ limit, however, the general trend, i.e., the positive correlation between $\sigma_{g,n}^2$ and $H_n$ should hold. To verify the relation between $\sigma^2_{g,n}$ and $H_n$, we have computed the gradient variance $\sigma^2_{g,n}$ and $H_n$ numerically for all the cases studied in this paper. As shown in Fig.~\ref{fig:sigmas}A, $\sigma^2_{g,n}$ depends on $H_n$ across the full range of $H_n$. To test the dependence of $\sigma_{g,n}$ on the scale of $\Delta w_{k,n}$, we made an overall scale change for $\Delta \boldsymbol{w}_k$: $\Delta \boldsymbol{w}_k\rightarrow s_w \Delta \boldsymbol{w}_k $ with a constant $s_w\in(0,1]$ for all sample-$k$, which is equivalent to a scaling change of $\Delta \boldsymbol{a}_k\rightarrow s_w \Delta \boldsymbol{a}_k$ with $s_w$ measuring the relative difference between activities for test and training samples. We have computed the sharpness spectra $\sigma_{g,n}$ for different values of $s_w$ and compared them with the the Hessian spectrum $H_n$. As shown in Fig.~\ref{fig:scaling} in the SI, the dependence of $\sigma_{g,n}$ on $H_n$ for different values of $s_w$ follows the same general trend (monotonically increasing) with an overall scaling factor that depends on $s_w$.  
Furthermore, in our previous work~\cite{feng2020neural}, we defined a flatness $F_n$ of loss landscape in direction-$n$ based on a threshold for the loss function. As shown in Fig.~\ref{fig:flatness_sigma_g} in SI, $\sigma_{g,n}$ is inversely correlated with $F_n$. 

Put together, all the evidence show that the first generalization determinant $\sigma_{g,n}$ measures the sharpness of the training loss landscape ($L(\boldsymbol{w})$) in direction-$n$ in the neighborhood of the solution (minimum) with the size of the neighborhood given by $\sigma_{w,n}$.

{\bf Geometrical interpretation of $\sigma_{w,n}$.} The meaning of the second generalization determinant $\sigma^2_{w,n}$ is more straightforward. It measures the variation of the dual weight change vector $\boldsymbol{\Delta w}_k$ projected onto direction-$n$ among all samples. From the expression of the minimal dual weights in Eq.~\ref{opt}, we have:
\begin{equation}
\label{sigmaW}
||\boldsymbol{\sigma_w}||^2\equiv \sum_{n=1}^M \sigma_{w,n}^2=\sum_{i=1}^{H_1}\sum_{i'=1}^{H_1}\sum_{j=1}^{H_2}C_{ii'} w_{ij}w_{i'j},
\end{equation}
where $C_{ii'}=\langle \frac{\Delta a_{i,k} \Delta a_{i',k}}{||\boldsymbol{a}_k||^2}\rangle_k$ is the matrix element of the covariance matrix $\boldsymbol{\mathcal{C}}$ of the relative difference $\boldsymbol{\Delta a}/||\boldsymbol{a}||_2$ of the neuron activities between training and testing samples. Note that $C_{ii'}$ depends only on the neuron activities and is independent of the weight $\boldsymbol{w}$ itself. 

From the quadratic form of Eq.~\ref{sigmaW}, it is clear that $||\boldsymbol{\sigma_w}||$ is a distance measure of the solution $\boldsymbol{w}$ weighted by the activity-dependent covariance matrix $C$. Indeed, as shown in Fig.~\ref{fig:sigmas}B, $||\boldsymbol{\sigma_w}||^2$ scales with the $L_2$-norm of the solution $||\boldsymbol{w}||_2$ in all the cases studied in this paper, which suggests that the second generalization determinant $\sigma_{w,n}$ represents a activity-weighted distance of the solution to the origin ($\boldsymbol{w}=\boldsymbol{0}$) along direction-$n$. We thus call $\sigma_{w,n}$ the size of the solution along direction-$n$. However, in addition to the weight solution $\boldsymbol{w}$, $\sigma_{w,n}$ also depends strongly on the activity difference between the testing and training data $\Delta\boldsymbol{a}$ via the covariance matrix $\boldsymbol{\mathcal{C}}$: when $\Delta\boldsymbol{a}\rightarrow 0$, $\sigma_{w.n}\rightarrow 0$.

\subsection{Understanding generalization through the lens of the two geometric determinants}

The dependence of the generalization gap on the two geometrical determinants as revealed in Eq.~\ref{GG} indicates two general strategies to improve generalization by seeking solutions with smaller $\sigma_{g,n}$ or smaller $\sigma_{w,n}$ or both (see Fig.~\ref{fig:lr2} in SI for an illustration of the two strategies).
 
In this section, we show how some of the well-known algorithms and regularization schemes for improving generalization follow exactly these two strategies (finding flatter or smaller solutions) or combination of both.

We first train neural network models to reach different solutions by using different hyperparameters (batch size and learning rate) in stochastic gradient descent (SGD), or by using explicit regularization schemes such as weight decay with different decay rates. 
For each solution, we computed the values of the two standard deviations ($\sigma_{g,n}$ and $\sigma_{w,n}$) in each eigen-direction $n$, and the correlation constant $c_n$. The generalization gap is decomposed into the contributions from different eigen-directions according to Eq.~\ref{GG}. 

Consistent with previous studies~\cite{feng2020neural,sagun2017empirical,yang2022does}, there are only a few sharp directions with large values of $\sigma_{g,n}$, and the loss landscape in most eigendirections is flat with much smaller values of $\sigma_{g,n}$. 
To quantify the contributions from the sharp directions and the flat directions, we separate the generalization gap into two parts: $\Delta L =\Delta L_s +\Delta L_f$ with $\Delta L_s=\sum_{n=1}^{n_s}\Delta L_n$ and $\Delta L_f=\sum_{n=n_s+1}^{N}\Delta L_n$ corresponding to the generalization gap from the sharp and flat directions, respectively, where $n$ is the rank order according to $\sigma_{g,n}$ and $n_s$ is the number of the sharp directions, which is defined by the value of $n$ with the steepest decrease in $\sigma_{g,n}$ (see Fig.~\ref{fig:sharp_peak} in SI). Despite their small number, the few sharp directions contribute to a significant fraction of the total generalization gap, e.g., in the example shown in Fig.~\ref{fig:B_alpha}, roughly $50\%$ of the generalization gap comes from the $n_s=10$ sharpest directions in the $900$-dimensional weight space. However, the contribution from the flat directions $\Delta L_f$ is not negligible due to the large number of flat directions.  

In previously proposed sharpness-based measures~\cite{KeskarMNST17,jiang2019fantastic} for generalization, only the sharpest direction(s) was considered. As we show below, one of the main insights gained from our study is that both $\Delta L_s$ and $\Delta L_f$ contribute significantly to the overall generalization gap. However, they can have different dependence on various hyper-parameters and generalization schemes as they depend on the two generalization determinants ($\sigma_{g,n}$ and $\sigma_{w,n}$) differently. As a result, they can be regularized and controlled independently or together to improve generalization.

\subsubsection{Finding flatter solutions by tuning SGD hyperparameters}

\begin{figure}[htbp]
\centering
\includegraphics[width=0.9\linewidth]{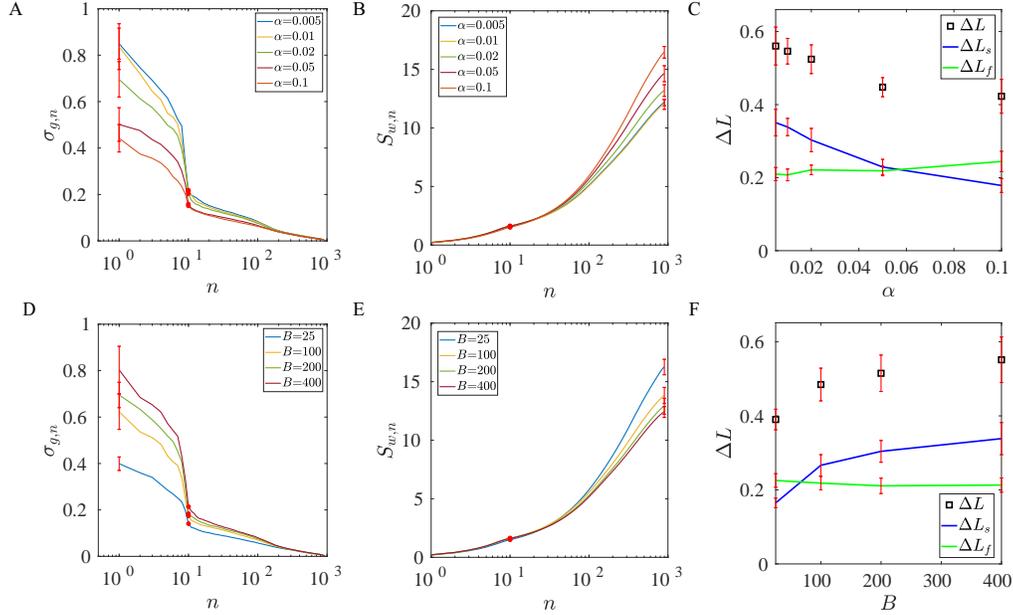}
\caption{The effects of the learning rate $\alpha$ or the batch size $B$ in SGD on generalization. (A) The sharpness spectrum $\sigma_{g,n}$ versus $n$ for different learning rates. (B) The accumulative size $S_{w,n}=\sum_{i=1}^n \sigma_{w,i}$ for different learning rates. (C) The generalization gap ($\Delta L$) and the contributions from the sharp and flat directions ($\Delta L_s$ and $\Delta L_f$) versus $\alpha$. (D-F) The same as (A-C) but for varying batch size $B$. All components are ordered by the decreasing order of $\sigma_{g,n}$ (from sharp directions to flat directions). As shown in (A) and (D), when increasing $\alpha$ or decreasing $B$, the sharpness ($\sigma_{g,n}$) decreases in the dominant components ($n\le n_s$) while $S_{w,n}$ does not change significantly as shown in (B)\&(E). As a result, the decrease in generalization gap comes mainly from the contribution from the sharp directions $\Delta L_s$ (blue lines) as shown in (C)\&(F).
Each line is averaged over $10$ independent realization and the error bar represents the standard deviation. $n_s=10$ (red dots) in this study. $B=25$ in (A-C) and $\alpha = 0.1$ in (D-F). }
\label{fig:B_alpha}
\end{figure}

We first change the learning rate $\alpha$ in SGD (with a fixed batch size) to obtain different solutions. We then analyze the solutions in terms of the generalization gap and its dependence on loss function sharpness ($\boldsymbol{\sigma}_{g}$) and solution size ($\boldsymbol{\sigma}_{w}$), see Methods for details.  
In Fig.~\ref{fig:B_alpha}A, we plotted the ``sharpness" spectrum, i.e., $\sigma_{g,n}$ versus $n$ for different values of $\alpha$, which clearly shows that as $\alpha$ increases, the values of $\sigma_{g,n}$ in the sharpest directions ($n\le n_s$) are reduced while they do not change significantly in the flatter directions ($n>n_s$). However, as shown in Fig.~\ref{fig:B_alpha}B, changing $\alpha$ does not significantly affect $\sigma_{w,n}$ across all the directions (Note that we plotted the accumulative sum $S_{w,n}\equiv \sum_{i=1}^n \sigma_{w,i}$ to smooth out the noise in $\sigma_{w,n}$). 
As a result, the generalization gap $\Delta L$ decreases when the learning rate increases from $0.005$ to $0.1$ ( further increase of $\alpha$ leads to non-convergence). More importantly,   
we find that the improvement in generalization for larger $\alpha$ is mainly due to the reduction of $\Delta L_s$ in the sharp directions while the contribution from the flat directions $\Delta L_f$ remains unchanged, which can be explained by the dependence of the two generalization determinants on $\alpha$ shown in Fig.~\ref{fig:B_alpha}A\&C. 

We next changed batch size $B$ (with a fixed $\alpha$) to obtain different solutions. By carrying out the same analysis as for changing $\alpha$, we found that increasing $B$ has the opposite effects as increasing $\alpha$. As shown in Fig.~\ref{fig:B_alpha}D-F, a larger $B$ leads to solutions where the sharpness parameters $\sigma_{g,n}$ in the sharpest directions increases, which leads to a larger generalization gap that originates from the increase in the contribution $\Delta L_s$ from the sharpest directions. The same dependence on $B$ and $\alpha$ is observed in different FCLs of the network (see Fig.~\ref{fig:multilayer_B} in SI for the case of varying $B$).

Previous work\cite{zhu2019anisotropic,Chaudhari_2018,feng2020neural} showed that the anisotropic landscape-dependent SGD noise drives the system away from sharp minima. Since the strength of the SGD noise scales with $\alpha/B$, the preference for flat minima is stronger for larger $\alpha$ and/or smaller $B$, which explains the improved generalization as $\alpha$ increases (or $B$ decreases) shown in Fig.~\ref{fig:B_alpha}. In fact, the dependence of generalization on $\alpha$ and $B$ collapsed onto a single dependence of $\Delta L$ on $\alpha/B$ as shown in Fig.~\ref{fig:flatness_alpha_B} in the SI.

\subsubsection{Smaller solutions are more generalizable: effects of weight decay and initialization}

\begin{figure}[htbp]
\centering
\includegraphics[width=0.9\linewidth]{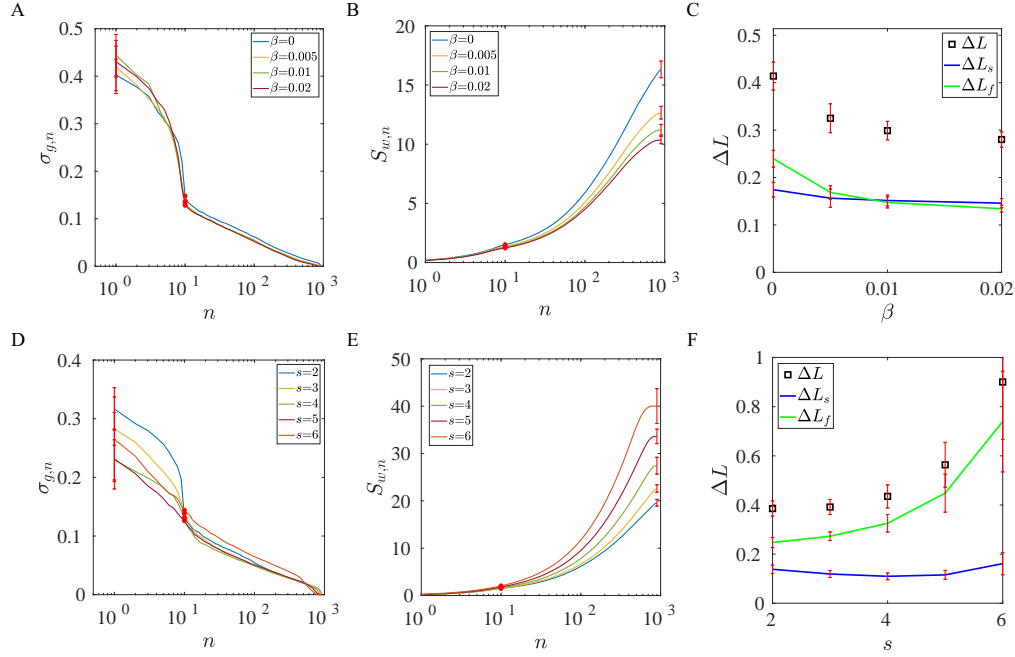}
\caption{The effects of the weight decay rate $\beta$ or the weight initiation $s$ in SGD on generalization. (A) The sharpness spectrum $\sigma_{g,n}$ versus $n$ for different values of $\beta$. (B) The accumulative size  $S_{w,n}=\sum_{i=1}^n \sigma_{w,i}$ for different values of $\beta$. (C) The generalization gap ($\Delta L$) and the contributions from the sharp and flat directions ($\Delta L_s$ and $\Delta L_f$) versus $\beta$. (D-F) The same as (A-C) but for different values of $s$. All components are ordered by the decreasing order of $\sigma_{g,n}$ (from sharp directions to flat directions). As shown in (A) and (D), when increasing $\beta$ or decreasing $s$, the sharpness ($\sigma_{g,n}$) does not change significantly while $S_{w,n}$ decreases as shown in (B)\&(E). As a result, the decrease in generalization gap comes mainly from the reduction of the loss from the flat directions $\Delta L_f$ (green lines) as shown in (C)\&(F). $B=25$ and $\alpha=0.1$ are used. }
\label{fig:beta_s}
\end{figure}

Another important regularization scheme to enhance generalization is by introducing weight decay in the learning dynamics. The weight decay scheme is equivalent to adding a regularization term that is proportional to the $L_2$ norm of the weight vector, $\beta ||\boldsymbol{w}||^2$, where the coefficient $\beta$ is a hyperparameter corresponding to the decay rate of weights. As shown in Fig.~\ref{fig:beta_s}A\&B, the sharpness of the solution does not depend on $\beta$ while the size of the solution characterized by the accumulative sum $S_{w,n}\equiv \sum_{i=1}^n \sigma_{w,i}$ is smaller when $\beta$ increases. As a result, the generalization gap $\Delta L$ decreases as the decay rate $\beta$ increases from $0$ to $0.02$, similar to the case when the learning rate $\alpha$ is increased. However, in contrast to the case of increasing learning rate $\alpha$, $\Delta L_s$ does not change significantly with $\beta$ and the reduction of $\Delta L$ comes mostly from the decrease of $\Delta L_f$ with increasing $\beta$, as shown in Fig.~\ref{fig:beta_s}C. 
The reduction in $\sigma_{w,n}$ is stronger in the flat directions since the resistance to weight reduction (decay) is weaker in flatter directions (see Fig.~\ref{fig:ratio_sigma_w} in SI), which explains the significant reduction in $\Delta L_f$ but not in $\Delta L_s$. Thus, our analysis shows that the enhancement of generalization by using weight decay is achieved predominantly by reducing the size of the solution, i.e., the second determinant of generalization. The same dependence on $\beta$ is observed in different layers of the network (see Fig.~\ref{fig:multilayer_beta} in SI).

Our approach can also be used to study the effect of timing in applying regularization such as weight decay~\cite{golatkar2019time}. In particular, we have trained our system by stopping weight decay at various time $t_\beta$. After $t_{\beta}$, we continue training the network without weigh decay until the training loss reach the required threshold. As shown in Fig.~\ref{fig:t_beta} in the SI, $t_\beta$ does not affect the sharpness spectrum $\sigma_{g,n}$ significantly (Fig.~\ref{fig:t_beta}A), and the size of the solution as characterized by $S_{w,n}$ decreases with increasing $t_\beta$ (Fig.~\ref{fig:t_beta}B). However, the decrease of $S_{w,n}$ slow down for $t_\beta>100~epoch$ and $S_{w,n}$ does not change significantly after $t_\beta$ reaches certain threshold $t^*_\beta \sim 200~epoch$. As a result, the generalization gap saturates after $t_\beta\ge 200~epoch$ as shown in Fig.~\ref{fig:t_beta}C, which is consistent with previous work~\cite{golatkar2019time}. The intuitive reason for this behavior is that although initially weight decay can decrease the size of the solution, after certain time $t^*_\beta$, the two forces for weight dynamics, i.e., the weight decay and the loss function gradient are balanced, and the weights does not change significantly anymore. 

To further elucidate the effect of solution size on generalization, we look for solutions of different sizes by using different initializations. In our experiment, the weights are initialized by Xavier initialization where weights are drawn randomly from a uniform distribution with a hyperparameter $s$ that determines the range of the uniform distribution. As shown in Fig.~\ref{fig:beta_s}D-F, the increase of $s$ has the opposite effects as compared with increasing $\beta$. As $s$ increases, the network finds solutions with similar sharpness spectra $\sigma_{g,n}$ but an increasing size as measured by $S_{w,n}$ shown in Fig.~\ref{fig:beta_s}. This leads to an increase in the generalization gap $\Delta L$, which comes mostly from the increase of the contribution from the flat directions $\Delta L_f$. 
These results confirm that a smaller solution with the same sharpness has a higher generalizability as shown by our theory.

\subsection{A case study: Enhancing generalization with decentralized multiple learners}
To demonstrate the utility of our theoretical results, we present a case study on enhancing generalization in the mutiple-learner setting. Due to massive computational requirements, training of realistic large deep nets is carried out by a distributed deep learning (DDL) approach with multiple processors (learners) where each learner computes the gradient for a different minibatch at each iteration and gradients from all learners are combined for updating the weights. The degree of parallelism in a DDL system is dictated by the total batch size (summed over learners): the larger the batch size, the more parallelism and higher speedup can be expected. Here, we use our A-W duality based analysis framework to determine which algorithm in DDL setting leads to solutions with better generalization and the reason behind it. 

First, we briefly describe the learning dynamics of the SGD-based algorithms with multiple ($m >1$) learners indexed by $j=1,2,3,...m$. At time (iteration) $t$, each learner has its own weight vector $\boldsymbol{W}_j(t)$ with their average given as: 
\begin{equation}
\boldsymbol{W}_a(t)\equiv m^{-1}\sum_{j=1}^{m} \boldsymbol{W}_j(t).
\end{equation}
Each learner $j$ updates its weights according to the cross-entropy loss function $L^{\mu_j(t)}(\vec{w})$ for minibatch $\mu_j(t)$ assigned to it at time $t$. The size of the local minibatch is $B$, and the overall batch size for all learners is $mB$. Two multi-learner algorithms are described below. 

(1) Synchronous Stochastic Gradient Descent (SSGD): In the SSGD algorithm, each learner $j\in[1,m]$ starts from the average weight vector $\boldsymbol{W}_a$ and moves along the gradient of its local loss function  $L^{\mu_j(t)}$ evaluated at the average weight $\boldsymbol{W}_a$:
\begin{equation}
\boldsymbol{W}_j(t+1)=\boldsymbol{W}_a(t)-\alpha \nabla L^{\mu_j(t)}(\boldsymbol{W}_a(t)),
\end{equation}
where $\alpha$ is the learning rate. It is easy to see that SSGD is equivalent to a single learner SGD with a large batch size $nB$. SSGD is the de facto DDL algorithm widely used in every DL application domain.

(2) Decentralized Parallel SGD (DPSGD): DPSGD is a state-of-the-art DDL algorithm that is proved to have similar convergence rate as SSGD and it runs much faster than SSGD algorithm when network latency is high and/or computational devices run at different speed \cite{dpsgd, adpsgd}. In the DPSGD algorithm \cite{dpsgd}, each learner $j$ computes the gradient at its own local weight $\boldsymbol{W}_j(t)$. The learning dynamics follows:
\begin{equation}
\boldsymbol{W}_j(t+1)=\boldsymbol{W}_{s,j}(t)-\alpha \nabla L^{\mu_j(t)}(\boldsymbol{W}_j(t)).
\end{equation}
where $\boldsymbol{W}_{s,j}(t)$ is the starting weight set to be the average weight of a subset of ``neighboring" learners of learner-$j$, 
e.g., $\boldsymbol{W}_{s,j}=\boldsymbol{W}_a$ if all learners are included as neighbors and $\boldsymbol{W}_{s,j}=(\boldsymbol{W}_{j-1}+\boldsymbol{W}_{j+1})/2$ if only two adjacent learners are included.

\begin{figure}[htbp]
\centering
\includegraphics[width=0.9\linewidth]{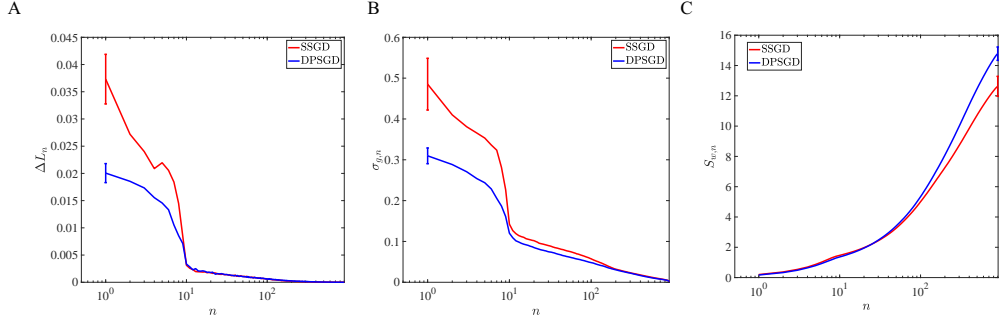}
\caption{The generalization gap and the two generalization determinants in different direction-$n$. (A) $\Delta L_n$ versus $n$ for SSGD (blue) and DPSGD (red). (B) $\sigma_{g,n}$ versus $n$. (C) $S_{w,n}$ versus $n$.}
\label{fig:DPSGD}
\end{figure}

To gain insights on generalization in the multiple learner setting, we first studied and compared the generalization performance for SSGD and DPSGD with the small MNIST dataset using a simple neural network model with 2 fully connected hidden layers. We used $m = 5$ and $B=200$ and the nearest-neighbor average $\boldsymbol{W}_{s,j}=(\boldsymbol{W}_{j-1}+\boldsymbol{W}_{j+1})/2$ for learner-$j$ was used in DPSGD. 
With the large overall batch size ($mB=1000$), we found that the DPSGD solution has a lower test error (2.3$\%$) than that of the SSGD solution (2.6$\%$). We next computed the generalization gap ($\Delta L_n$), and the two determinants for generalization ($\sigma_{g,n}$ and $\sigma_{w,n}$) in each direction-$n$ for both SSGD and DPSGD solutions by following the analysis framework developed in the previous sections. As shown in Fig.~\ref{fig:DPSGD}, the generalization gap is much reduced for the DPSGD solution in the sharp directions ($n<10$), which is caused by the fact that the sharpness $\sigma_{g,n}$ is much reduced in the sharp directions ($n<10$), while the size of the solution ($\sigma_{w,n}$) remains unchanged. Thus, our analysis indicates that DPSGD has a better generalization performance than SSGD as it can find flatter solutions without increasing the size of the solution.    

\begin{figure}[htbp]
\centering
\includegraphics[width=0.7\linewidth]{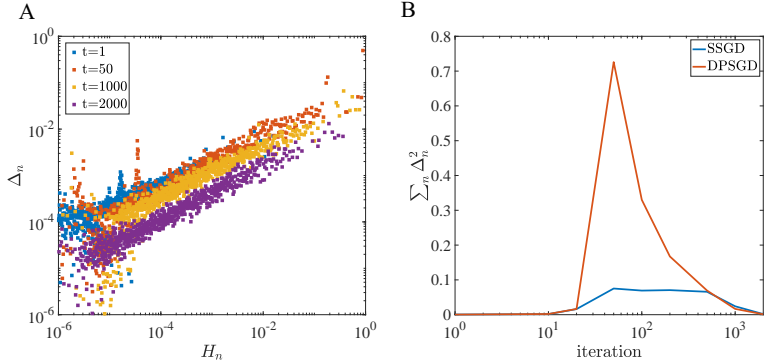}
\caption{(A) DPSGD noise strength $\Delta_n$ versus eigenvalue $H_n$ of the Hessian of the loss function in different directions ($n$) at different training times. $n$ represents the Hessian eigendirections ranked ordered by the Hessian eigenvalues.  (B) Noise strength of SSGD (blue) and DPSGD (blue). To compute the noise strength, we generate an ensemble of gradients of minibatch loss functions over an ensemble of random minibatches ($100$ minibatches were used here). Noise strength $\Delta_n$ is calculated as the standard deviation of the projections of these gradients onto direction-$n$.}
\label{fig:DPSGD_noise}
\end{figure}

Why can DPSGD find flatter solutions? It was recently shown that the landscape-dependent noise in SGD-based algorithms can drive the system towards flat minima~\cite{feng2020neural}. However, the noise in SSGD is inversely proportional to the total batch size ($mB$)~\cite{Chaudhari_2018,feng2020neural} and it becomes too small in the large batch setting. In DPSGD, however, different learners compute their gradients at their own weights $\boldsymbol{W}_j$ that are different from their mean $\boldsymbol{W}_a$. As a result, the difference in gradients $\delta \boldsymbol{g}_j\equiv \nabla L^{\mu_j}(\boldsymbol{W}_j)- \nabla L^{\mu_j}(\boldsymbol{W}_a)$ leads to an additional noise term in DPSGD. By expanding $\delta g_j$ w.r.t. a small $\delta \boldsymbol{W}_j\equiv \boldsymbol{W}_j-\boldsymbol{W}_a$, we have $\delta \boldsymbol{g}_j =\boldsymbol{\mathcal{H}}^{\mu_j} \delta \boldsymbol{W}_j$ where $\boldsymbol{\mathcal{H}}^{\mu_j}=\nabla\nabla L^{\mu_j}$ is the Hessian of the minibatch loss function $L^{\mu_j}$. This suggests that the additional DPSGD noise depends on the loss landscape and it is larger when the loss landscape is sharper, which has been verified by direct simulations as shown in Fig.~\ref{fig:DPSGD_noise}A. Note that SSGD noise also has the same qualitative dependence on the loss landscape as shown in previous study~\cite{feng2020neural}, but its strength is smaller than the DPSGD noise due to the additional landscape-dependent noise introduced by DPSGD especially during the training time window before the system reaches its final solution when the landscape is relatively rough as shown in Fig.~\ref{fig:DPSGD_noise}B. 

\begin{figure}[htbp]
\centering
\includegraphics[width=0.9\linewidth]{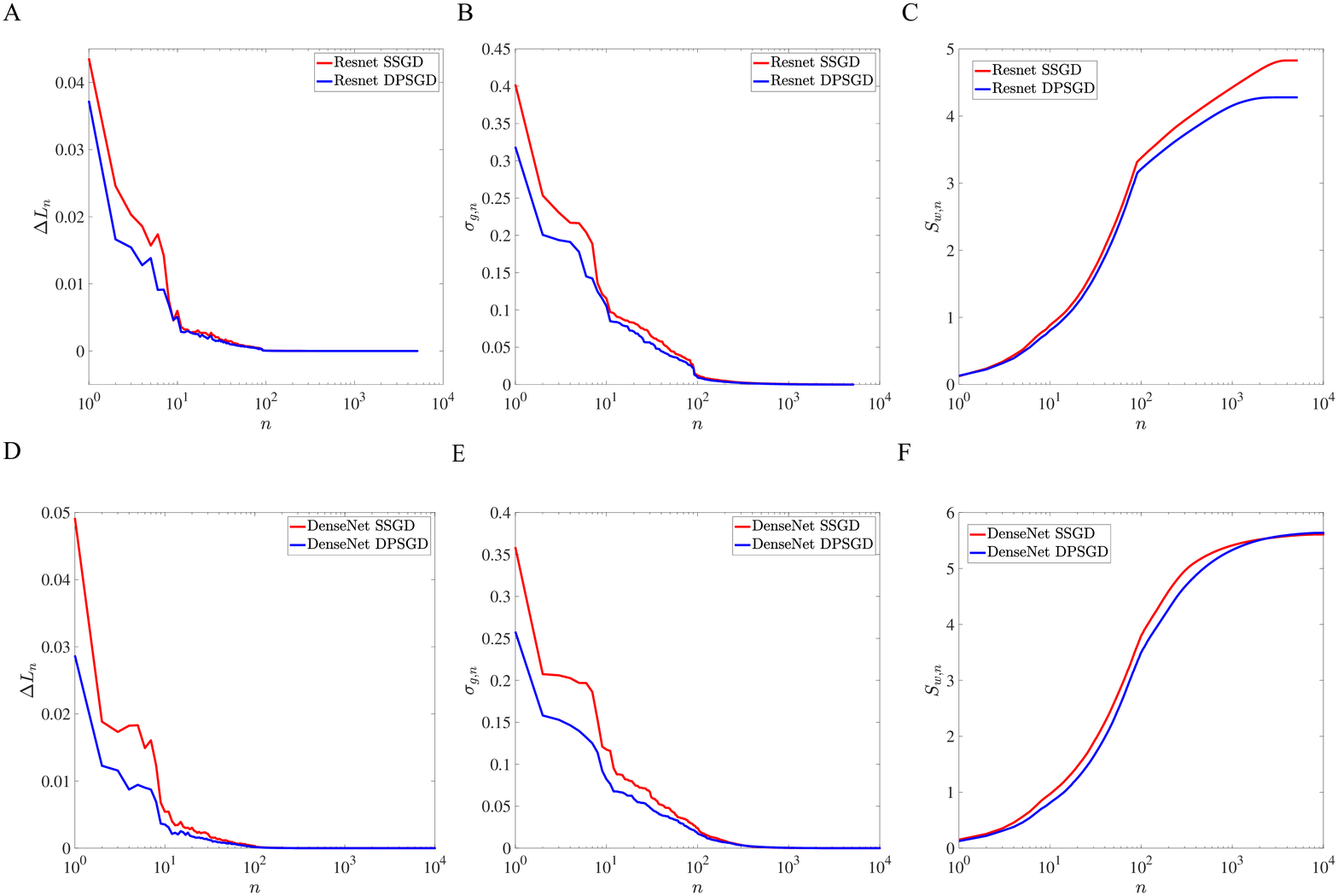}
\caption{Analysis of generalization for DPSGD and SSGD in large models (Resnet and Densenet). (A) $\Delta L_n$ versus $n$ for Resnet. (B) $\sigma_{g,n}$ versus $n$ for Resnet. (C) $S_{w,n}$ versus $n$ for Resnet. (D) $\Delta L_n$ versus $n$ for Densenet. (E) $\sigma_{g,n}$ versus $n$ for Densenet. (F) $S_{w,n}$ versus $n$ for Densenet. The analysis was done for the last layer of the network models. See Methods for details of the Resnet and Densenet models.}
\label{fig:resnet_densenet}
\end{figure}

The advantages of DPSGD in the large batch size setting for state-of-the-art large models with large datasets have been  demonstrated empirically in our previous study~\cite{dpsgd-large-batch} (see Methods section for a partial list). For example, for the full CIFAR-10 dataset, DPSGD achieves a higher test accuracy ($94.34\%$) than SSGD ($92.70\%$) for the Resnet-18 model~\cite{resnet}; and DPSGD ($94.79\%$) also outperforms SSGD ($92.79\%$) for the Dense-121 model~\cite{densenet} (see Methods for details of the \resnet and \densenet models used here). Here, to understand the underlying reason for their different generalization performances, we have computed the dual weights in the last layers of the \resnet and \densenet models, which are fully connected. From the dual weights, we calculated the two generalization determinants ($\sigma_{g,n}$ and $\sigma_{w,n}$) and the generalization gap decomposition $\Delta L_n$ in different eigendirection-$n$ of the Hessian (with respect to the weights in the last layer). As shown in Fig.~\ref{fig:resnet_densenet}, for both models (\resnet and \densenet), in comparison with the SSGD solution, the sharpness $\sigma_{g,n}$ for the DPSGD solution is smaller in the sharp directions ($n<10$), while the size of the solution ($\sigma_{w,n}$) remains approximately the same. As a result, the generalization gap for the DPSGD solution is smaller in the sharp directions, which leads to a better generalization performance for the DPSGD solution. 

Overall, from our analysis of generalization performance based on the activity-weight duality for both simple and more realistic large neural network models, we show that DPSGD has a better generalization performance than SSGD as it can find flatter solutions without increasing the size of the solution. Furthermore, the ability of finding flatter solutions in DPSGD is due to the additional landscape dependent noise introduced by the different weights at which different learners compute their gradients.

\section{Summary and Discussion}

In this work, we discovered an exact duality (equivalence) relation between changes in the activities of the neurons in a densely connected layer of neurons and the changes of the weights connecting this layer to the next. By using the duality relation, we decompose the generalization gap into contributions from different directions in the weight space and the contribution from each direction-$n$ depends on the product of the sharpness of the loss landscape ($\sigma_{g,n}$) and the standard deviation of the dual weights ($\sigma_{w,n}$), which represent an activity-weighted distance of the solution to the origin. These two factors, $\sigma_{g,n}$ and $\sigma_{w,n}$, describe the geometric properties of the loss landscape at the solution and the solution itself, respectively. While $\sigma_{g,n}$ depends predominantly on the training loss landscape, $\sigma_{w,n}$ depends on the relative difference between testing and training data. Together, these two geometric factors determine the generalization gap.

As far as we know, the notion that flat minima correspond to more generalizable solutions was first put forth by Hochreiter and Schmidhuber in 1997~\cite{hochreiter1997flat}. Empirical evidence in DLNN were found in support of this idea~\cite{KeskarMNST17,wei2021improved}, which has been used to develop algorithms searching for flatter minima~\cite{chaudhari2016entropysgd,foret2021sharpnessaware,Baldassi2020shaping} to improve generalization. However, as first pointed out by Dinh et al~\cite{dinh2017sharp}, the loss function of a feed forward network is invariant under a simple scaling transformation in which the weights in one layer are multiplied by a common scaling factor $p$ and the weights in the next layer is scaled by $p^{-1}$. This scale invariance means that one can always find a solution whose sharpness in a given layer is arbitrarily large without affecting its generalization gap, which casts serious doubt on using sharpness of the loss function as the sole measure for generalization.

Indeed, flatness has the dimension of weight, which is different from that of the loss function. Therefore, even on a purely dimensional ground, flatness can not be the only determinant for generalization. The key question is that when we say the loss landscape is flat, what is it compared with? Here, we show that the flatness of the loss landscape in a given direction-$n$ should be compared with the variation of the dual weights in that direction characterized by $\sigma_{w,n}$, which is the second determinant for generalization. This second determinant is related to the weight norm, which is another popular measure of generalization~\cite{neyshabur2015normbased,bartlett2017spectrallynormalized,neyshabur2018understanding,nagarajan2019generalization}. However, weight norm alone has not been a successful measure of generalization in DLNN~\cite{jiang2019fantastic}. One of the main insights gained from our study is that these two determinants together determine generalization. For the scale transformation proposed by Dinh et al~\cite{dinh2017sharp}, the sharpness ($\boldsymbol{\sigma}_g$) in a given layer scales by $p$, and the weight variation ($\boldsymbol{\sigma}_w$) in the same layer scales by $p^{-1}$, which makes the expression for generalization gap (Eq.~\ref{GG}) scale invariant as it should be.  

Although the primary goal of our study is not to set bounds for the generalization gap, our results shed some lights on how such a bound would depend on. One insight gained from our study is that contributions ($\Delta L_s$ and $\Delta L_f$) from the sharp and the flat directions are both important for the overall generalization gap. The contribution from the few sharp directions ($n_c\ll M$) is significant due to their large sharpness $\sigma_{g,n}$. However, since the number of flat directions is much larger than $n_c$, the total contribution from the flat directions $\Delta L_f$ is comparable with $\Delta_s$. Both $\Delta L_s$ and $\Delta L_f$ depend on the sharpness and the size of the solution but in different ways. 
More specifically, $\Delta L_s$ is strongly affected by changes in $\sigma_{g,n}$, which occur mostly in the sharp directions ($n\le n_c$), while $\Delta L_f$ is affected predominantly by the size of the solution $\sigma_{w,n}$. Indeed, in a recent study by Yang et al~\cite{yang2022does}, the authors obtained an analytical non-vacuous PAC-Bayes bound with a similar decomposition of the generalization gap into the contributions from the sharp and the flat (``sloppy") directions as well as a contribution that depends on the weight norm. By using appropriate Gaussian prior and posterior that have sloppy spectra with separation of sharp and flat modes, the authors achieved a tight bound for the generalization gap (within a factor of two for LeNet on MNIST), which is far more superior than other PAC-Bayes bounds. 

The A-W duality and the resulting decomposition of the generalization gap can be obtained in any densely connected layer in a neural network model. 
Most neural network models no matter how complex they are contain FCLs, where we can apply our analysis to understand and compare generalizability of different solutions. This makes our method versatile as we demonstrated for different FCLs in the simple multilayer perceptron models for the MNIST dataset as well as for the more complex \resnet and \densenet models for the CIFAR-10 dataset. Furthermore, given the relatively small number of weights in a given FCL, the A-W duality based analysis can be done efficiently.    
For a convolution layer, there is no exact duality relation due to the relatively small number of weights for each filter. However, the general idea of looking for equivalence between activity and weight changes may be worth pursuing. For example, one may obtain an approximate duality relation, e.g., by minimizing the difference of the output activities $||\boldsymbol{\phi}(\boldsymbol{a},\boldsymbol{w})-\boldsymbol{\phi}(\boldsymbol{a'},\boldsymbol{w'})||$. If such an approximate duality relation leads to a good approximation of the overall loss function, we expect the general conclusions reached in this paper may still be valid.

In this paper, we used SGD and weight decay as two representative examples to highlight the dependence of the generalization gap on either one of the two determinants, i.e., the sharpness of the loss landscape at the solution and the size of the solution. In general, both determinants can be affected, albeit in different ways depending on the regularization scheme used and specifics of the data set, see section 2 in SI for the analysis on effects of dropout, data size and mislabeled data on generalization. Going forward, the insights gained from our analysis of the key determinants for generalization based on the A-W duality may provide general guidance in developing new algorithms and/or regularization schemes to improve generalization as well as understanding possible scaling dependence of generalization on the number of parameters and the size of training data in the large model and large data size limit. The A-W duality, which connects changes in activities with changes in weights, may also be useful for understanding the connection between sloppy spectra in the data space and in the solution (weight) space~\cite{yang2022does} as well as for studying robust learning against noise in data and adversarial attack.   

\section{Methods}

\textbf{Data Set and Neural Network Architecture.} We used a subset of MNIST dataset or CIFAR-10 dataset as our training data. The subset contains all ten classes, with $N$ training images per class. The size of test data is 1,000-2,000 images for both case. 

For MNIST experiment, we trained two fully connected networks with multiple hidden layers where $l$ th hidden layer contains $H_l$ hidden units. In the main text, we did the experiments on a network with two hidden layers, and each hidden layer contains 30 hidden units ($H_1 = H_2 = 30$). In the SI, we showed the experimental results of a network with four hidden layers, where each layer has the same number of hidden units ($H_1=H_2=H_3=H_4=30$). 

For the CIFAR-10 dataset, we used a convolutional neural network with two convolutional layers and four fully connected layers. The size of two convolutional layers are $3 \times 5 \times 5 \times 6$ and $6 \times 5 \times 5 \times 16$, where the size is denoted by number of input channels $\times$ kernel size $\times$ kernel size $\times$ number of output channels. Each convolutional layer is followed by a max-pooling layer with size $2\times2$. The size of four fully connected layers are 400, 120, 25, 10.

\textbf{Simulation Details.} Unless stated otherwise, the default hyperparameters and data size are: $\alpha = 0.1, B = 25, \beta = 0, s = 1, N = 400, \rho = 0, d= 0$. Stochastic gradient descent (SGD) was used for training.

The weights are initialized by Xavier initialization: weights are drawn from a uniform distribution $U(-\frac{s\sqrt{6}}{\sqrt{n_i + n_{i+1}}},\frac{s\sqrt{6}}{\sqrt{n_i + n_{i+1}}})$, where $n_i$ is the number of incoming connections to the layer and $n_{i+1}$ is the number of ongoing connections from the layer. 

During training, a weight-decay regularization with decay rate $\beta$ is used for the first 200 epochs. The weights are considered to be a solution when its corresponding training error reaches 0. In practical, we stop the training when training loss first reached a low threshold = $5 \times 10^{-4}$ for MNIST experiments and $1 \times 10^{-3}$ for CIFAR-10 experiments.

\textbf{State-of-the-art neural network models and large-scale experiments.} To better verify our theory in the real-world scenario, we chose two state-of-the-art models for the CIFAR-10 tasks: \resnet, a 18 layer instantiation of ResNet architecture \cite{resnet} and \densenet, a 121 layer instantiation of DenseNet architecture \cite{densenet}. In both \resnet and \densenet, each convolution layer has input channel dimension 3, output channel dimension 64 and kernel size of 3. The linear layer of \resnet is of shape 512 $\times$ 10 and the linear layer of \densenet is of shape 1024 $\times$ 10. In both \resnet and \densenet, we use Batch-Norm layer to reduce covariance shift. We used batch size 512 per GPU and a total of 16 v100 GPUs to train both tasks. We used SGD optimizer with momentum 0.9, learning rate 3.2 (annealed by 10x at the 160-th epoch and again at the 240-th epoch). We trained 320 epochs for both tasks. SSGD took 0.35 hour (\resnet) and 0.68 hour (\densenet) over 16 v100 GPUs.  In contrast, DPSGD took 0.32 hour (\resnet) and 0.58 hour (\densenet). 

Furthermore, our previous empirical study\cite{dpsgd-large-batch}   agrees with our theory in that we  found that DPSGD introduces additional landscape-dependent noise that automatically adjusts the effective learning rate to improve convergence. The results are consistent across different large-scale application domains: Computer Vision (CIFAR10 and ImageNet-1K), Automatic Speech Recognition (SWB300 and SWB2000) and Natural Language Processing (Wikitext-103); three different types of neural network models: Convolutional Neural Networks, Long Short-Term Memory Recurrent Neural Networks and Attention-based Transformer Models; and two optimizers: SGD and Adam.

\textbf{Analysis Details.} Once the network is trained to the predefined accuracy, we analyze the solution in the following way:
\begin{itemize}
    \item[1.] For each test data, we calculate its Euclidean distance with all training data and find its closest partner.
    \item[2.] Choose two layers with full connection and calculate the dual weight for each test data by using the activity-weight duality relation. For each pair of training data and test data, we first calculate the neural activities $x_i$ for the chosen layer and obtain the neural activity difference $\Delta x_i$ between training data and test data. With $x_i$ and $\Delta x_i$, the minimal duality solution is obtained by applying Eq.~\ref{opt}. In MNIST experiments, we did the activity-weight duality analysis between two hidden layers; In CIAFR-10 experiments, we did the analysis between two fully connected hidden layers with size $120 \times 25$.
    \item[3.] The dual weight $w^*$ and the effective gradient $g$ were determined from Eqs.~\ref{opt}\&\ref{g_k}, respectively. The two vectors are projected onto the eigen-directions of Hessian matrix. The standard deviations of the $n$-th components, $\sigma_{g,n}$ and $\sigma_{w,n}$, as well as the correlation coefficient $c_n$ between them were computed. The  generalization gap was computed by using Eq.~\ref{GG}. 
\end{itemize}

\section{Data and Code Availability}
The paper uses data from MNIST and CIFAR-10. The code is avaliable at the public repository https://github.com/YuFengDuke/A-W-Duality-Project.

\section{Acknowledgments}
We thank Ken Clarkson and Roger Traub for careful reading of our manuscript and useful comments. The work by FY was partially done while he was employed as an IBM intern.

\bibliography{ML}

\begin{thebibliography}{10}
\expandafter\ifx\csname url\endcsname\relax
  \def\url#1{\texttt{#1}}\fi
\expandafter\ifx\csname urlprefix\endcsname\relax\def\urlprefix{URL }\fi
\providecommand{\bibinfo}[2]{#2}
\providecommand{\eprint}[2][]{\url{#2}}

\bibitem{LeCun2015Deep}
\bibinfo{author}{LeCun, Y.}, \bibinfo{author}{Bengio, Y.} \&
  \bibinfo{author}{Hinton, G.}
\newblock \bibinfo{title}{Deep learning}.
\newblock \emph{\bibinfo{journal}{Nature}} \textbf{\bibinfo{volume}{521}},
  \bibinfo{pages}{436 EP --} (\bibinfo{year}{2015}).
\newblock \urlprefix\url{https://doi.org/10.1038/nature14539}.

\bibitem{goodfellow2016deep}
\bibinfo{author}{Goodfellow, I.}, \bibinfo{author}{Courville, A.} \&
  \bibinfo{author}{Bengio, Y.}
\newblock \emph{\bibinfo{title}{Deep learning}}, vol.~\bibinfo{volume}{1}
  (\bibinfo{publisher}{MIT Press}, \bibinfo{year}{2016}).

\bibitem{he2016deep}
\bibinfo{author}{He, K.}, \bibinfo{author}{Zhang, X.}, \bibinfo{author}{Ren,
  S.} \& \bibinfo{author}{Sun, J.}
\newblock \bibinfo{title}{Deep residual learning for image recognition}.
\newblock In \emph{\bibinfo{booktitle}{Proceedings of the IEEE conference on
  computer vision and pattern recognition}}, \bibinfo{pages}{770--778}
  (\bibinfo{year}{2016}).

\bibitem{wu2016google}
\bibinfo{author}{Wu, Y.} \emph{et~al.}
\newblock \bibinfo{title}{Google's neural machine translation system: Bridging
  the gap between human and machine translation}.
\newblock \emph{\bibinfo{journal}{arXiv preprint arXiv:1609.08144}}
  (\bibinfo{year}{2016}).

\bibitem{alpha-go}
\bibinfo{author}{Silver, D.} \emph{et~al.}
\newblock \bibinfo{title}{Mastering the game of go with deep neural networks
  and tree search}.
\newblock \emph{\bibinfo{journal}{Nature}} \textbf{\bibinfo{volume}{529}},
  \bibinfo{pages}{484--489} (\bibinfo{year}{2016}).
\newblock \urlprefix\url{https://doi.org/10.1038/nature16961}.

\bibitem{AlphaFold}
\bibinfo{author}{Jumper, J.} \emph{et~al.}
\newblock \bibinfo{title}{Highly accurate protein structure prediction with
  alphafold}.
\newblock \emph{\bibinfo{journal}{Nature}} \textbf{\bibinfo{volume}{596}},
  \bibinfo{pages}{583--589} (\bibinfo{year}{2021}).
\newblock \urlprefix\url{https://doi.org/10.1038/s41586-021-03819-2}.

\bibitem{zhang2016understanding}
\bibinfo{author}{Zhang, C.}, \bibinfo{author}{Bengio, S.},
  \bibinfo{author}{Hardt, M.}, \bibinfo{author}{Recht, B.} \&
  \bibinfo{author}{Vinyals, O.}
\newblock \bibinfo{title}{Understanding deep learning requires rethinking
  generalization}.
\newblock \emph{\bibinfo{journal}{ICLR}}  (\bibinfo{year}{2017}).

\bibitem{jiang2019fantastic}
\bibinfo{author}{Jiang, Y.}, \bibinfo{author}{Neyshabur, B.},
  \bibinfo{author}{Mobahi, H.}, \bibinfo{author}{Krishnan, D.} \&
  \bibinfo{author}{Bengio, S.}
\newblock \bibinfo{title}{Fantastic generalization measures and where to find
  them}.
\newblock \emph{\bibinfo{journal}{ICLR}}  (\bibinfo{year}{2020}).

\bibitem{KeskarMNST17}
\bibinfo{author}{Keskar, N.~S.}, \bibinfo{author}{Mudigere, D.},
  \bibinfo{author}{Nocedal, J.}, \bibinfo{author}{Smelyanskiy, M.} \&
  \bibinfo{author}{Tang, P. T.~P.}
\newblock \bibinfo{title}{On large-batch training for deep learning:
  Generalization gap and sharp minima}.
\newblock \emph{\bibinfo{journal}{ICLR}}  (\bibinfo{year}{2017}).

\bibitem{dinh2017sharp}
\bibinfo{author}{Dinh, L.}, \bibinfo{author}{Pascanu, R.},
  \bibinfo{author}{Bengio, S.} \& \bibinfo{author}{Bengio, Y.}
\newblock \bibinfo{title}{Sharp minima can generalize for deep nets}.
\newblock \emph{\bibinfo{journal}{Proceedings of the 34th International
  Conference on Machine Learning}} \textbf{\bibinfo{volume}{70}},
  \bibinfo{pages}{1019--1028} (\bibinfo{year}{2017}).

\bibitem{zhu2019anisotropic}
\bibinfo{author}{Zhu, Z.}, \bibinfo{author}{Wu, J.}, \bibinfo{author}{Yu, B.},
  \bibinfo{author}{Wu, L.} \& \bibinfo{author}{Ma, J.}
\newblock \bibinfo{title}{The anisotropic noise in stochastic gradient descent:
  Its behavior of escaping from sharp minima and regularization effects}.
\newblock In \emph{\bibinfo{booktitle}{Proc. Int. Conf. Mach. Learn.}},
  \bibinfo{pages}{7654--7663} (\bibinfo{year}{2019}).

\bibitem{martens2014new}
\bibinfo{author}{Martens, J.}
\newblock \bibinfo{title}{New insights and perspectives on the natural gradient
  method}.
\newblock \emph{\bibinfo{journal}{Journal of Machine Learning Research}}
  \textbf{\bibinfo{volume}{21}}, \bibinfo{pages}{1--76} (\bibinfo{year}{2020}).

\bibitem{feng2020neural}
\bibinfo{author}{Feng, Y.} \& \bibinfo{author}{Tu, Y.}
\newblock \bibinfo{title}{The inverse variance--flatness relation in stochastic
  gradient descent is critical for finding flat minima}.
\newblock \emph{\bibinfo{journal}{Proceedings of the National Academy of
  Sciences}} \textbf{\bibinfo{volume}{118}} (\bibinfo{year}{2021}).

\bibitem{sagun2017empirical}
\bibinfo{author}{Sagun, L.}, \bibinfo{author}{Evci, U.},
  \bibinfo{author}{Guney, V.~U.}, \bibinfo{author}{Dauphin, Y.} \&
  \bibinfo{author}{Bottou, L.}
\newblock \bibinfo{title}{Empirical analysis of the hessian of
  over-parametrized neural networks} (\bibinfo{year}{2017}).
\newblock \eprint{1706.04454}.

\bibitem{yang2022does}
\bibinfo{author}{Yang, R.}, \bibinfo{author}{Mao, J.} \&
  \bibinfo{author}{Chaudhari, P.}
\newblock \bibinfo{title}{Does the data induce capacity control in deep
  learning?}
\newblock In \emph{\bibinfo{booktitle}{International Conference on Machine
  Learning}}, \bibinfo{pages}{25166--25197} (\bibinfo{organization}{PMLR},
  \bibinfo{year}{2022}).

\bibitem{Chaudhari_2018}
\bibinfo{author}{Chaudhari, P.} \& \bibinfo{author}{Soatto, S.}
\newblock \bibinfo{title}{Stochastic gradient descent performs variational
  inference, converges to limit cycles for deep networks}.
\newblock \emph{\bibinfo{journal}{2018 Information Theory and Applications
  Workshop (ITA)}}  (\bibinfo{year}{2018}).
\newblock \urlprefix\url{http://dx.doi.org/10.1109/ita.2018.8503224}.

\bibitem{golatkar2019time}
\bibinfo{author}{Golatkar, A.~S.}, \bibinfo{author}{Achille, A.} \&
  \bibinfo{author}{Soatto, S.}
\newblock \bibinfo{title}{Time matters in regularizing deep networks: Weight
  decay and data augmentation affect early learning dynamics, matter little
  near convergence}.
\newblock \emph{\bibinfo{journal}{Advances in Neural Information Processing
  Systems}} \textbf{\bibinfo{volume}{32}} (\bibinfo{year}{2019}).

\bibitem{dpsgd}
\bibinfo{author}{Lian, X.} \emph{et~al.}
\newblock \bibinfo{title}{Can decentralized algorithms outperform centralized
  algorithms? a case study for decentralized parallel stochastic gradient
  descent}.
\newblock In \emph{\bibinfo{booktitle}{Advances in Neural Information
  Processing Systems}}, \bibinfo{pages}{5330--5340} (\bibinfo{year}{2017}).

\bibitem{adpsgd}
\bibinfo{author}{Lian, X.}, \bibinfo{author}{Zhang, W.},
  \bibinfo{author}{Zhang, C.} \& \bibinfo{author}{Liu, J.}
\newblock \bibinfo{title}{Asynchronous decentralized parallel stochastic
  gradient descent}.
\newblock In \emph{\bibinfo{booktitle}{ICML}} (\bibinfo{year}{2018}).

\bibitem{dpsgd-large-batch}
\bibinfo{author}{Zhang, W.} \emph{et~al.}
\newblock \bibinfo{title}{Loss landscape dependent self-adjusting learning
  rates in decentralized stochastic gradient descent}.
\newblock \emph{\bibinfo{journal}{CoRR}}
  \textbf{\bibinfo{volume}{abs/2112.01433}} (\bibinfo{year}{2021}).
\newblock \urlprefix\url{https://arxiv.org/abs/2112.01433}.
\newblock \eprint{2112.01433}.

\bibitem{resnet}
\bibinfo{author}{He, K.}, \bibinfo{author}{Zhang, X.}, \bibinfo{author}{Ren,
  S.} \& \bibinfo{author}{Sun, J.}
\newblock \bibinfo{title}{Deep residual learning for image recognition}.
\newblock \emph{\bibinfo{journal}{CVPR}}  (\bibinfo{year}{2015}).

\bibitem{densenet}
\bibinfo{author}{{Huang}, G.}, \bibinfo{author}{{Liu}, Z.},
  \bibinfo{author}{{Van Der Maaten}, L.} \& \bibinfo{author}{{Weinberger},
  K.~Q.}
\newblock \bibinfo{title}{Densely connected convolutional networks}.
\newblock In \emph{\bibinfo{booktitle}{2017 IEEE Conference on Computer Vision
  and Pattern Recognition (CVPR)}}, \bibinfo{pages}{2261--2269}
  (\bibinfo{year}{2017}).

\bibitem{hochreiter1997flat}
\bibinfo{author}{Hochreiter, S.} \& \bibinfo{author}{Schmidhuber, J.}
\newblock \bibinfo{title}{Flat minima}.
\newblock \emph{\bibinfo{journal}{Neural Computation}}
  \textbf{\bibinfo{volume}{9}}, \bibinfo{pages}{1--42} (\bibinfo{year}{1997}).

\bibitem{wei2021improved}
\bibinfo{author}{Wei, C.} \& \bibinfo{author}{Ma, T.}
\newblock \bibinfo{title}{Improved sample complexities for deep networks and
  robust classification via an all-layer margin}.
\newblock \emph{\bibinfo{journal}{ICLR}}  (\bibinfo{year}{2020}).

\bibitem{chaudhari2016entropysgd}
\bibinfo{author}{Chaudhari, P.} \emph{et~al.}
\newblock \bibinfo{title}{Entropy-sgd: Biasing gradient descent into wide
  valleys}.
\newblock \emph{\bibinfo{journal}{ICLR}}  (\bibinfo{year}{2017}).

\bibitem{foret2021sharpnessaware}
\bibinfo{author}{Foret, P.}, \bibinfo{author}{Kleiner, A.},
  \bibinfo{author}{Mobahi, H.} \& \bibinfo{author}{Neyshabur, B.}
\newblock \bibinfo{title}{Sharpness-aware minimization for efficiently
  improving generalization}.
\newblock \emph{\bibinfo{journal}{ICLR}}  (\bibinfo{year}{2021}).

\bibitem{Baldassi2020shaping}
\bibinfo{author}{Baldassi, C.}, \bibinfo{author}{Pittorino, F.} \&
  \bibinfo{author}{Zecchina, R.}
\newblock \bibinfo{title}{Shaping the learning landscape in neural networks
  around wide flat minima}.
\newblock \emph{\bibinfo{journal}{Proceedings of the National Academy of
  Sciences}} \textbf{\bibinfo{volume}{117}}, \bibinfo{pages}{161--170}
  (\bibinfo{year}{2020}).
\newblock \urlprefix\url{https://www.pnas.org/content/117/1/161}.
\newblock \eprint{https://www.pnas.org/content/117/1/161.full.pdf}.

\bibitem{neyshabur2015normbased}
\bibinfo{author}{Neyshabur, B.}, \bibinfo{author}{Tomioka, R.} \&
  \bibinfo{author}{Srebro, N.}
\newblock \bibinfo{title}{Norm-based capacity control in neural networks}.
\newblock \emph{\bibinfo{journal}{PMLR}} \textbf{\bibinfo{volume}{40}},
  \bibinfo{pages}{1376--1401} (\bibinfo{year}{2015}).

\bibitem{bartlett2017spectrallynormalized}
\bibinfo{author}{Bartlett, P.}, \bibinfo{author}{Foster, D.~J.} \&
  \bibinfo{author}{Telgarsky, M.}
\newblock \bibinfo{title}{Spectrally-normalized margin bounds for neural
  networks}.
\newblock \emph{\bibinfo{journal}{NeurIPS}} \bibinfo{pages}{6241--6250}
  (\bibinfo{year}{2017}).

\bibitem{neyshabur2018understanding}
\bibinfo{author}{Neyshabur, B.}, \bibinfo{author}{Li, Z.},
  \bibinfo{author}{Bhojanapalli, S.}, \bibinfo{author}{LeCun, Y.} \&
  \bibinfo{author}{Srebro, N.}
\newblock \bibinfo{title}{Towards understanding the role of
  over-parametrization in generalization of neural networks}.
\newblock \emph{\bibinfo{journal}{ICLR}}  (\bibinfo{year}{2019}).

\bibitem{nagarajan2019generalization}
\bibinfo{author}{Nagarajan, V.} \& \bibinfo{author}{Kolter, J.~Z.}
\newblock \bibinfo{title}{Generalization in deep networks: The role of distance
  from initialization}.
\newblock \emph{\bibinfo{journal}{NeurIPS}}  (\bibinfo{year}{2020}).

\bibitem{Srivastava2014DropoutAS}
\bibinfo{author}{Srivastava, N.}, \bibinfo{author}{Hinton, G.},
  \bibinfo{author}{Krizhevsky, A.}, \bibinfo{author}{Sutskever, I.} \&
  \bibinfo{author}{Salakhutdinov, R.}
\newblock \bibinfo{title}{Dropout: A simple way to prevent neural networks from
  overfitting}.
\newblock \emph{\bibinfo{journal}{Journal of Machine Learning Research}}
  \textbf{\bibinfo{volume}{15}}, \bibinfo{pages}{1929--1958}
  (\bibinfo{year}{2014}).

\bibitem{feng2021phases}
\bibinfo{author}{Feng, Y.} \& \bibinfo{author}{Tu, Y.}
\newblock \bibinfo{title}{Phases of learning dynamics in artificial neural
  networks in the absence or presence of mislabeled data}.
\newblock \emph{\bibinfo{journal}{Machine Learning: Science and Technology}}
  \textbf{\bibinfo{volume}{2}}, \bibinfo{pages}{043001} (\bibinfo{year}{2021}).

\end{thebibliography}

\newpage

\begin{center}

\section*{Supplemental Information}

\end{center}

\setcounter{equation}{0}
\setcounter{section}{0}

\renewcommand{\theequation}{S\arabic{equation}}
\setcounter{page}{1}

\renewcommand{\thepage}{SI.\arabic{page}}

\section{Leading order approximation for $\sigma_{g,n}$}
As described in the main text, the minimal dual weights $\boldsymbol{w}^*_k$ for test-train sample pair-$k$ can be expressed as: $\boldsymbol{w}^*_k=\boldsymbol{w}+\Delta\boldsymbol{w}_k$ where the components of $\Delta \boldsymbol{w}_k$ can be explicitly written as $\Delta w_{ij,k}=\sum_{i'=1}^{H_1} b_{i'i,k} w_{i'j,k}$ with $b_{i'i,k}= \frac{ \Delta a_{i',k} a_{i,k}}{||\boldsymbol{a}_k||^2}$. By flattening the weight matrix to form a weight vector with dimension $M$, we can write $\Delta \boldsymbol{w}_k=\sum_{n=1}^M \Delta w_{n,k}\boldsymbol{e}_n$ where $\boldsymbol{e}_n$ is the unit vector in the $n$-th direction in the weight space.

Following the procedure for defining partial derivatives in multi-dimensional space, we can write $\Delta l_k$ as the sum of differences between the functional values at two coordinates that differ only along one of the basis directions:
\begin{equation}
    \Delta l_k =\sum_{n=1}^M [l(\boldsymbol{x}_k,\boldsymbol{w}+\boldsymbol{\Delta \tilde{w}}_{n,k})-l(\boldsymbol{x}_k,\boldsymbol{w}+\boldsymbol{\Delta \tilde{w}}_{(n-1),k})],
\end{equation}
where $\boldsymbol{\Delta\tilde{w}}_{n,k}=(\Delta w_{1,k}, \Delta w_{2,k},...,\Delta w_{n,k},0,..,0)$ is the projection of $\boldsymbol{\Delta w}_k$ onto the first $n$ directions.  Note that $\boldsymbol{\Delta \tilde{w}}_{n,k} - \boldsymbol{\Delta\tilde{w}}_{(n-1),k}=(\boldsymbol{\Delta w} _k\cdot \boldsymbol{e}_n)\boldsymbol{e}_n=\Delta w_{n,k}\boldsymbol{e}_n$ is the projection of $\boldsymbol{\Delta w}_k$ onto direction-$n$ and we can define $d(n,k)\equiv l(\boldsymbol{x}_k,\boldsymbol{w}+\boldsymbol{\Delta \tilde{w}}_{n,k})-l(\boldsymbol{x}_k,\boldsymbol{w}+\boldsymbol{\Delta \tilde{w}}_{(n-1),k})]$ as the change of loss along direction-$n$. 

The overall generalization gap can be approximated as the average pair-wise generalization gap: $\Delta L\approx \Delta \tilde{L}=\sum_n \langle d(n,k)\rangle_k$. If we neglect the effect of correlation between different components of $\boldsymbol{\Delta w}$ on the loss function, we can approximate $\langle d(n,k)\rangle_k$ by taking all the unchanged weight components in $l(\boldsymbol{x}_k,\boldsymbol{w}+\boldsymbol{\Delta \tilde{w}}_{n,k})$ and $l(\boldsymbol{x}_k,\boldsymbol{w}+\boldsymbol{\Delta \tilde{w}}_{(n-1),k})$ to be their average values (over $k$):
\begin{equation}
\langle d(n,k)\rangle_k \approx \langle l( \boldsymbol{x}_k,\boldsymbol{w}+\Delta w_{n,k}\boldsymbol{e}_n) - l(\boldsymbol{x}_k,\boldsymbol{w})\rangle _k,  
\label{dnk}
\end{equation}
where we have used the fact that $\langle  \boldsymbol{\Delta w}_k\rangle _k=\boldsymbol{0}$. 

By using linear approximation, the expression in Eq.~\ref{dnk} can be further simplified to: $\langle d(n,k)\rangle _k \approx  g^{(0)}_{n,k} \Delta  w_{n,k}$ where $g^{(0)}_{n,k}=\frac{\partial l_k}{\partial w_n}|_w$ is the gradient of the loss function $l_k$ for sample-$k$ in direction-$n$. Thus, the generalization can be expressed as:
\begin{eqnarray}
    \Delta L &\approx& \langle l(\boldsymbol{x}_k,\boldsymbol{w}+\boldsymbol{\Delta w}_k) - l(\boldsymbol{x}_k,\boldsymbol{w})\rangle_k= \sum_{n=1}^M  \langle g_{n,k} \Delta w_{n,k}\rangle_k
    \\
    &\approx& \sum_{n=1}^N \langle l(\boldsymbol{x}_k,\boldsymbol{w}+\Delta w_{n,k}\boldsymbol{e}_n) - l(\boldsymbol{x}_k,\boldsymbol{w})\rangle_k \\
    &\approx& \sum_{n=1}^N \langle  g^{(0)}_{n,k} \Delta  w_{n,k}\rangle_k,\label{g0}
\end{eqnarray}
where the two successive approximations are made by ignoring higher order correlation effects and taking the linear approximation, respectively. Thus, the leading order approximation of $\sigma^2_{g,n}\approx \langle (g^{(0)}_{n,k})^2\rangle_k$, which is the diagonal elements of the expected Fisher information matrix $\boldsymbol{\mathcal{F}}=\langle \nabla_{\boldsymbol{w}} l_k\nabla_{\boldsymbol{w}} l_k\rangle _k$ at the solution $\boldsymbol{w}$ where $l_k=l(\boldsymbol{x}_k,\boldsymbol{w})$ is the training loss for sample-$k$. 

Next, we convert $l_k$ into a log-likelihood $q_k$ by making the following transformation:
\begin{equation}
q_k=Q^{-1}e^{l_k},
\end{equation}
where $Q=\sum_k e^{l_k}$ is the normalization factor so that $\sum_k q_k=1$. According to known results, the expected Fisher information matrix of the log-likelihood is equal to the expected Hessian matrix at the solution: 
\begin{equation}
\langle \nabla \ln(q_k) \nabla \ln(q_k) \rangle_k = \langle \nabla^2 \ln(q_k)\rangle _k.
\end{equation}
If we assume the individual loss is small: $\l_k\ll 1$, we can approximate $Q$ as: $Q\approx \sum_{k=1}^{N_{te}} (1+l_k)=N_{te}(1+L)$ where $L=\langle l_k\rangle_k $ is the average training loss. Plugging in the approximate expression $\ln(q_k)\approx l_k-\ln(1+L)-\ln N_{te}$ in the above equation and taking into account the fact that the gradient $\nabla L=0$ at the solution, we have:
\begin{equation}
\boldsymbol{\mathcal{F}}=\langle \nabla \l_k \nabla l_k \rangle_k \approx  \frac{L}{1+L} \nabla \nabla L\approx L\nabla \nabla L=L\boldsymbol{\mathcal{H}},
\label{Fisher_Hessian}
\end{equation}
where $\boldsymbol{\mathcal{H}}\equiv \nabla \nabla L$ is the Hessian matrix of the training loss $L$ at the solution. Note that we have used the approximation $L\ll 1$ in the above equation. Eq.~\ref{Fisher_Hessian} shows that the expected (empirical) Fisher information matrix of the loss function $l_k$ is linearly proportional to the Hessian of the average loss at the solution.  

\section{The effects of dropout, data size, and mislabeled data}

In the main text of the paper, we used SGD and weight decay as two representative examples to highlight the dependence of the generalization gap on either one of the two determinants, the sharpness of the loss landscape at the solution and the size of the solution, respectively. In general, both determinants can be affected. For example, another popular regularization scheme is dropout~\cite{Srivastava2014DropoutAS} where a randomly selected subset of neurons (and their connected weights) are dropped during each iteration. We find that both the sharpness of loss landscape and the size of the solution are affected by the dropout fraction $d$ but in opposite ways. As $d$ increases from zero, $\sigma_{g,n}$ in the sharpest directions first decrease significantly before saturating to fixed values while $\sigma_{w,n}$ continuously increase with $d$ (see Fig.~\ref{fig:dropout}A\&B). At low $d$, $\Delta L$ is dominated by $\Delta L_s$, which decreases with $d$. At high $d$, $\Delta L$ is dominated by $\Delta L_f$, which increases with $d$. As a result, the generalization gap has a non-monotonic dependence on $d$ as shown in Fig.~\ref{fig:dropout}C.

In addition to various regularization schemes, we have also studied the effects of the size ($N$) of the training data set. We found that both $\sigma_{g,n}$ and $\sigma_{w,n}$ change with $N$: the solution becomes flatter across all directions but also slightly larger as shown in Fig.~\ref{fig:N_rho}A-D. In addition, the correlation coefficient $c_n$ decreases with $N$ (see Fig.~\ref{fig:N_rho}C), which also contributes to reducing the generalization gap in all directions. The combined effect is that even though the reduction in $\Delta L$ comes largely from the reduction of $\Delta L_s$ from the sharp direction the decrease of $\Delta L_f$ with $N$ is not negligible, i.e., increasing $N$ improves generalization in all directions across the sharpness spectrum.

We have also investigated the case where a fraction ($\rho$) of the training data have random labels (see Fig.~\ref{fig:N_rho}E-H)~\cite{zhang2016understanding, feng2021phases}. As expected, the generalization gap ($\Delta L$) increases with $\rho$ (Fig.~\ref{fig:N_rho}H). As $\rho$ increases, the sharpness of all the flat directions ($n>n_s$) increases significant while the sharpness of the few sharpest directions ($n\le n_s)$ decreases (Fig.~\ref{fig:N_rho}E). For a finite $\rho$, the sharpness spectrum is continuous without a sudden drop seen in the case with $\rho=0$ -- the network needs to use all weight directions to memorize the random labels of the mislabeled data. Besides sharpness, the size of the solution also increases with $\rho$. In addition, the correlation coefficient $c_n$ increases with $\rho$ for all $n$. The combined effect is that $\Delta L$ is dominated by $\Delta L_f$, which increases with $\rho$ due to the increase of both the sharpness $\sigma_{g,n}$ and the size $\sigma_{w,n}$ in almost all the directions ($n>n_s$).

\newpage

\section*{Supporting figures}

\setcounter{figure}{0}
\renewcommand{\figurename}{Fig.}
\renewcommand{\thefigure}{S\arabic{figure}}

\begin{figure}[htbp]
\centering
\includegraphics[width=0.55\linewidth]{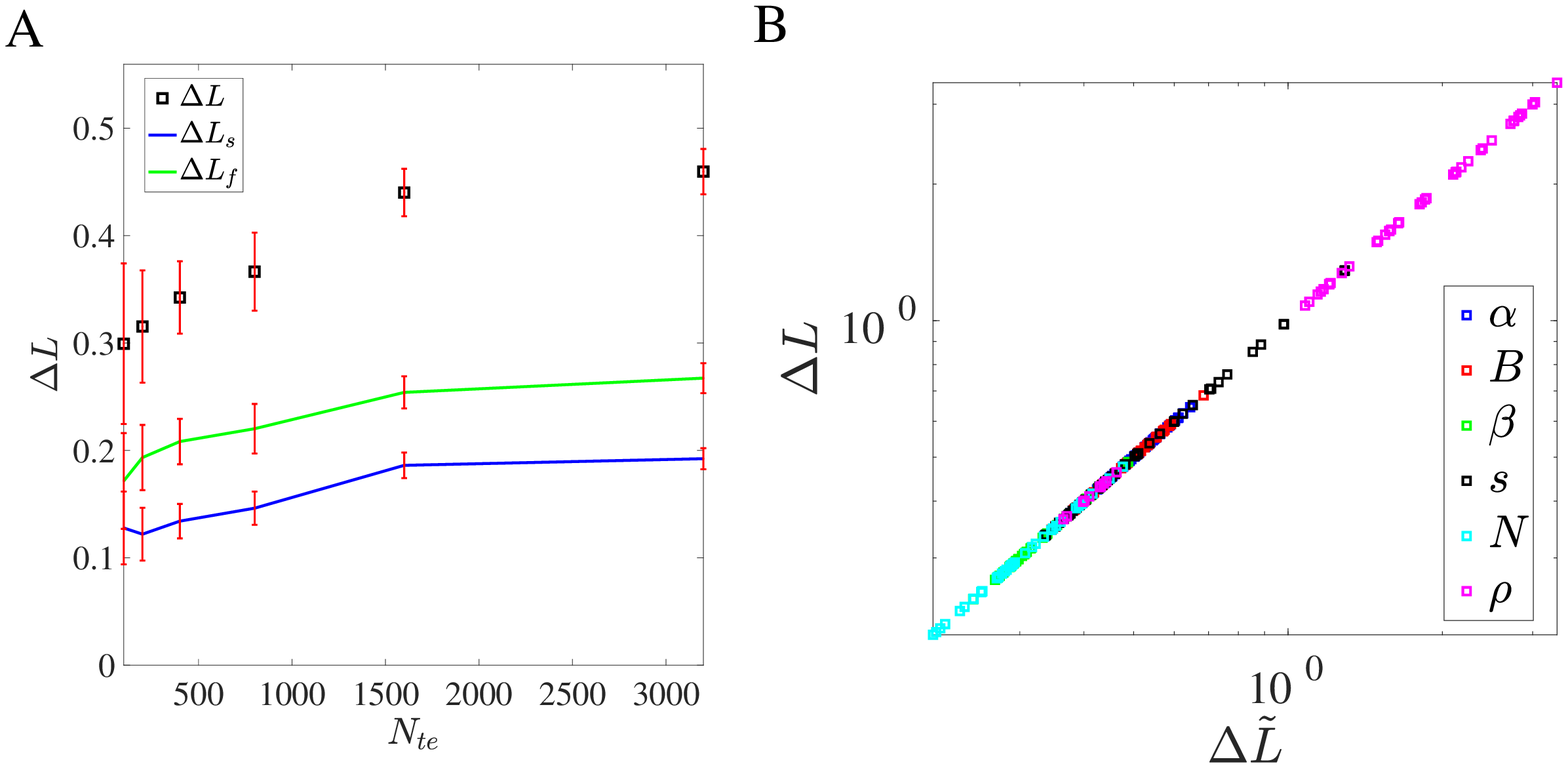}
\caption{The  dependence of $\Delta L$ on the number of test samples $N_{te}$ for the MNIST dataset. $\Delta L$ converges quickly when $N_{te}\ge 2,000$, which is much smaller than the overall size of the dataset. Thus, a small subset of test data is enough to carry out the duality analysis. }
\label{fig:pairloss}
\end{figure}

\begin{figure}[htbp]
\centering
\includegraphics[width=1\linewidth]{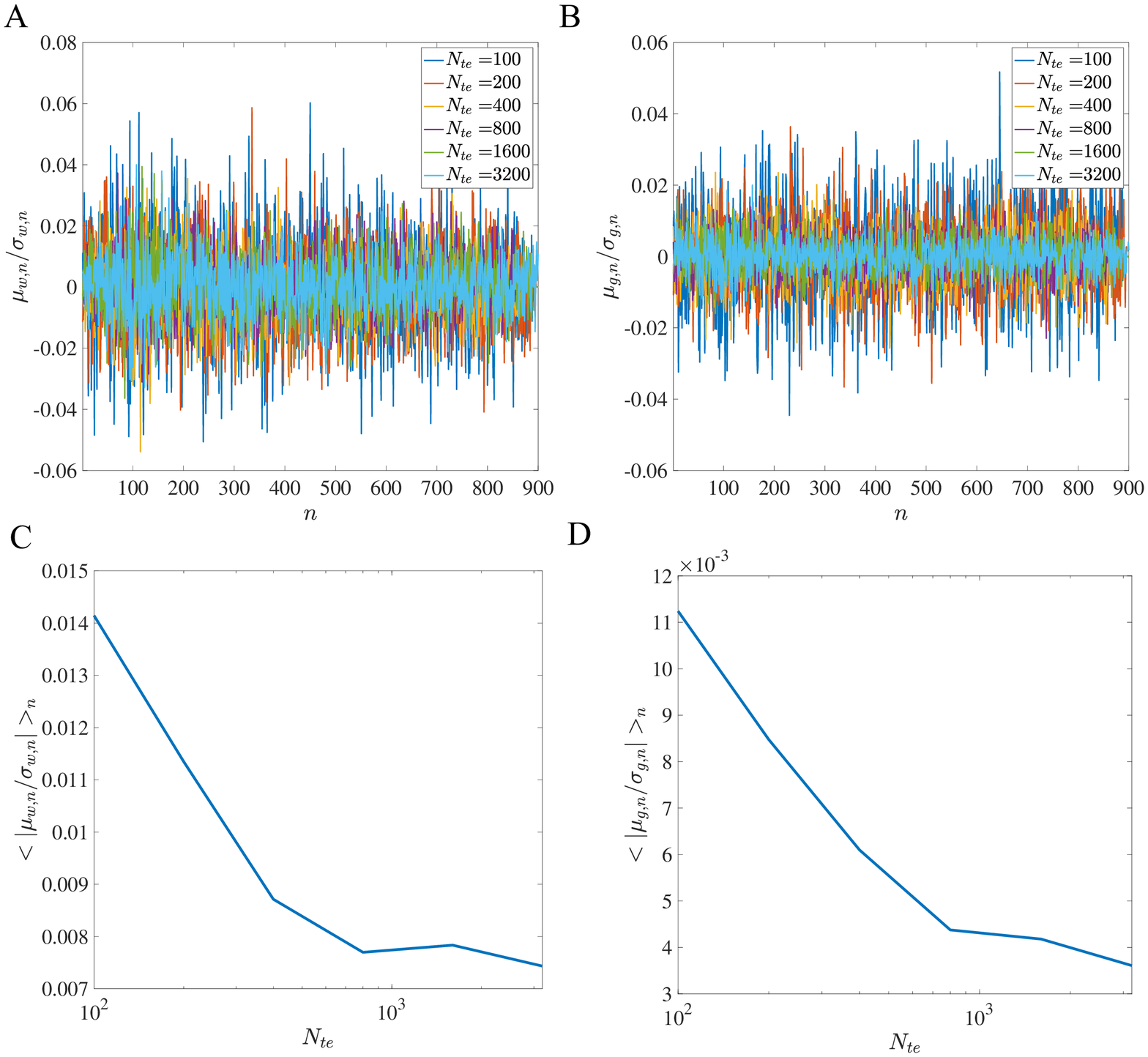}
\caption{The average components $\mu_{w,n}$ and $\mu_{g,n}$ relative to the corresponding standard deviations $\sigma_{w,n}$ and $\sigma_{g,n}$. (A) $\mu_{w,n}/\sigma_{w,n}$ versus $n$ for different values of $N_{te}$. (B) $\mu_{g,n}/\sigma_{g,n}$ versus $n$ for different values of $N_{te}$. (C) The dependence of the averaged ratio $\langle |\mu_{w,n}/\sigma_{w,n}|\rangle_n$ on $N_{te}$. (D) The dependence of the averaged ratio $\langle |\mu_{g,n}/\sigma_{g,n}|\rangle_n$ on $N_{te}$. From our results, it is clear that $|\mu_{w,n}|\ll \sigma_{w,n}$, $|\mu_{g,n}|\ll \sigma_{g,n}$ and the average ratios $\langle |\mu_{w,n}/\sigma_{w,n}|\rangle_n$ and $\langle |\mu_{g,n}/\sigma_{g,n}|\rangle_n$ both decrease with $N_{te}$ before saturating to small values that are much less than $1$. Results are averaged over $10$ independent realizations, hyperparameters used are $\alpha = 0.1, B = 25$. }
\label{fig:mu_g&w}
\end{figure}

\begin{figure}[htbp]
\centering
\includegraphics[width=1\linewidth]{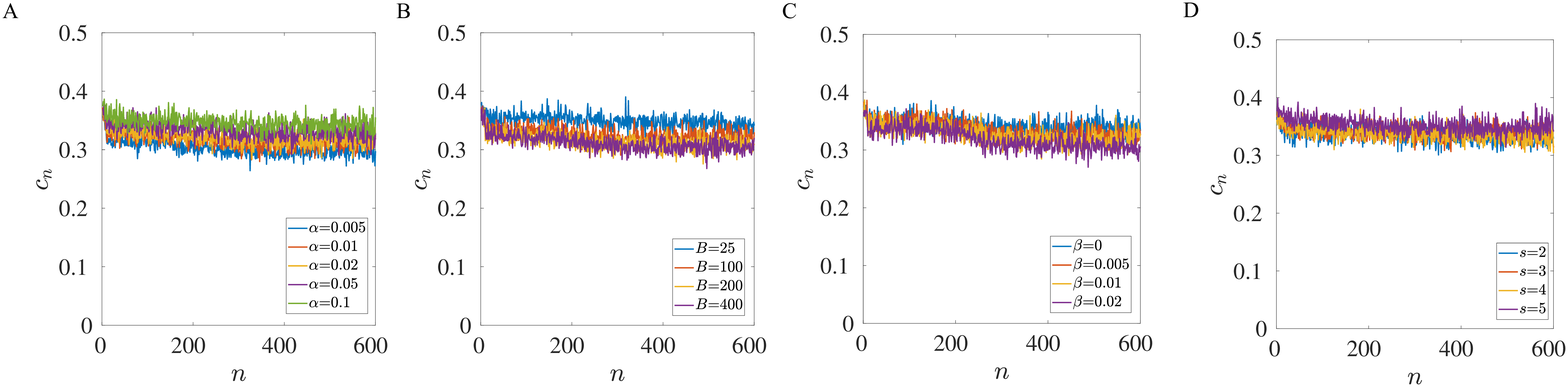}
\caption{The correlation coefficient $c_n$ for different hyperparameters. $c_n$ does not change significantly with $n$ except for very large $n$ where the contributions to the generalization gap is negligible. Except for adding new data (correctly labeled data or mislabeled data), $c_n$ remain roughly constant independent of the hyperparameters.}
\label{fig:c_n}
\end{figure}

\begin{figure}[htbp]
\centering
\includegraphics[width=1\linewidth]{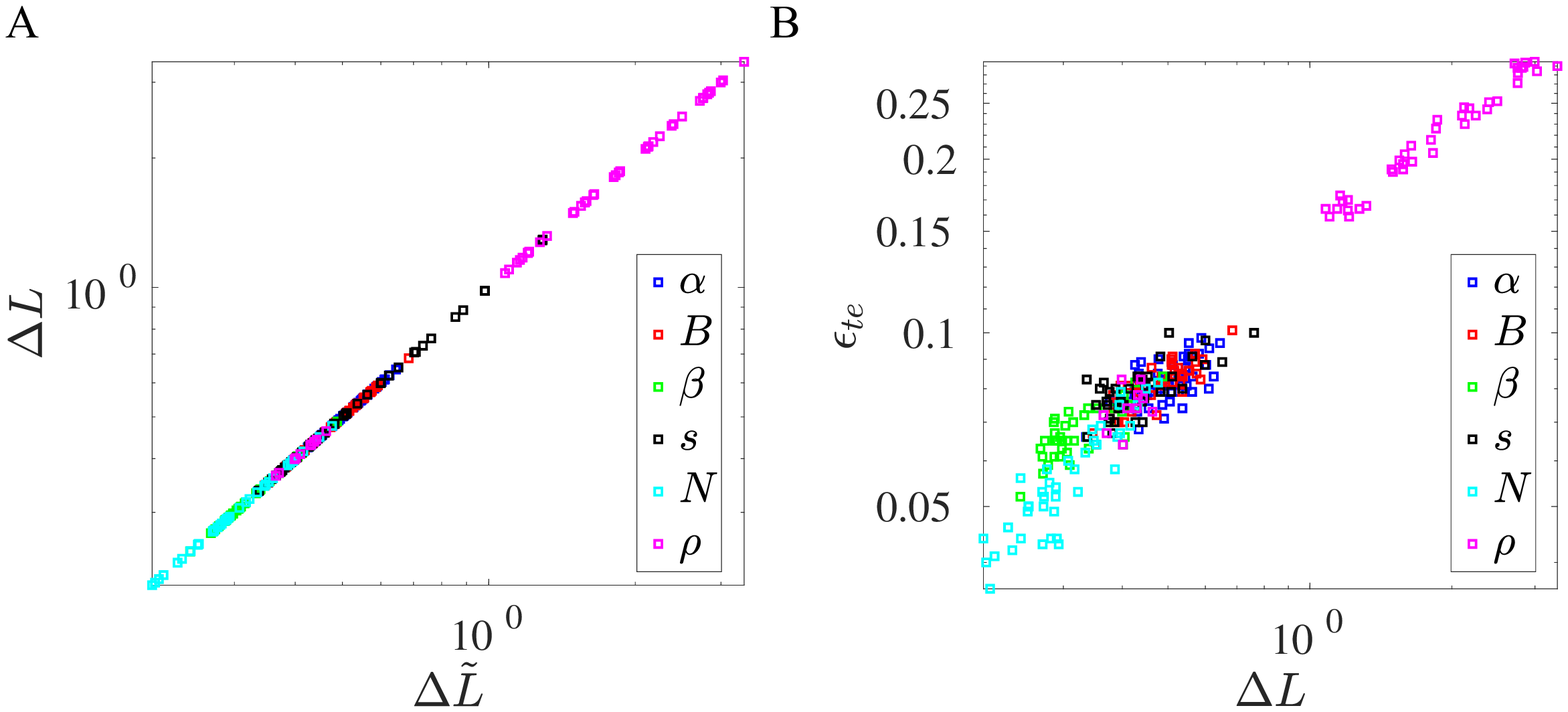}
\caption{(A) The pair-wise generalization gap $\Delta\tilde{L}=L'-\tilde{L}=\langle l(\boldsymbol{x}_k,\boldsymbol{w})-l(\boldsymbol{x'}_k,\boldsymbol{w})\rangle_k$ computed directly from data versus $\Delta L$ determined from Eq.~\ref{GG} using the A-W duality.
(B) The test error $\epsilon_{te}$ versus $\Delta L$ (note the training error $\epsilon_{tr}=0$ in the examples shown here).
}
\label{fig:test}
\end{figure}

\begin{figure}[htbp]
\centering
\includegraphics[width=1\linewidth]{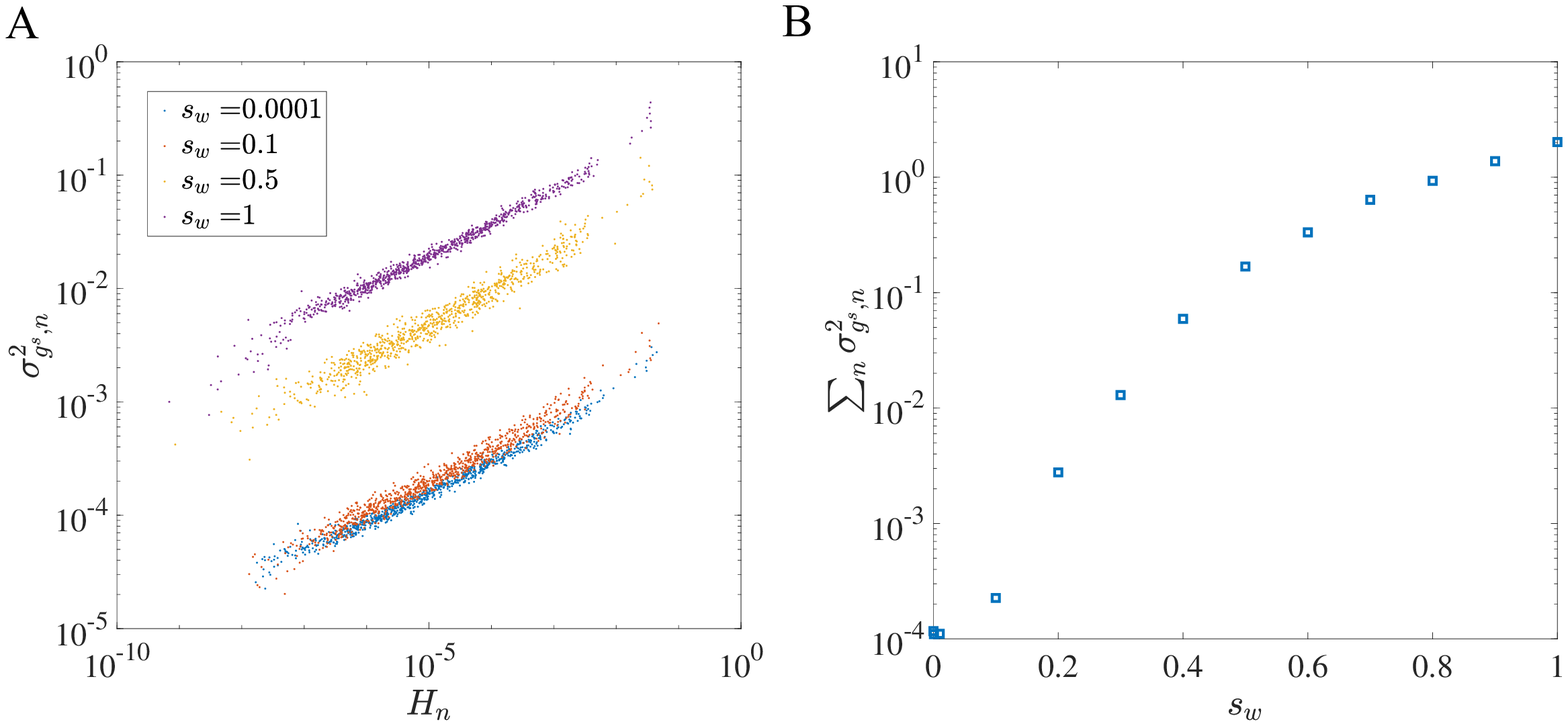}
\caption{The dependence on the scaling factor $s_w$ for the MNIST dataset. (A) The dependence of $\sigma_{g,n}^2$ on $H_n$ for different values of $s_w$. All the curves follow the same trend. (B) The total variance $\sum_n \sigma_{g,n}^2$ increases with $s_w$. $s_w$ is used to scale $\Delta \boldsymbol{w}_k$: $\Delta \boldsymbol{w}_k\rightarrow s_w \Delta \boldsymbol{w}_k $ (see main text for details). }
\label{fig:scaling}
\end{figure}
\FloatBarrier

\begin{figure}[htbp]
\centering
\includegraphics[width=0.55\linewidth]{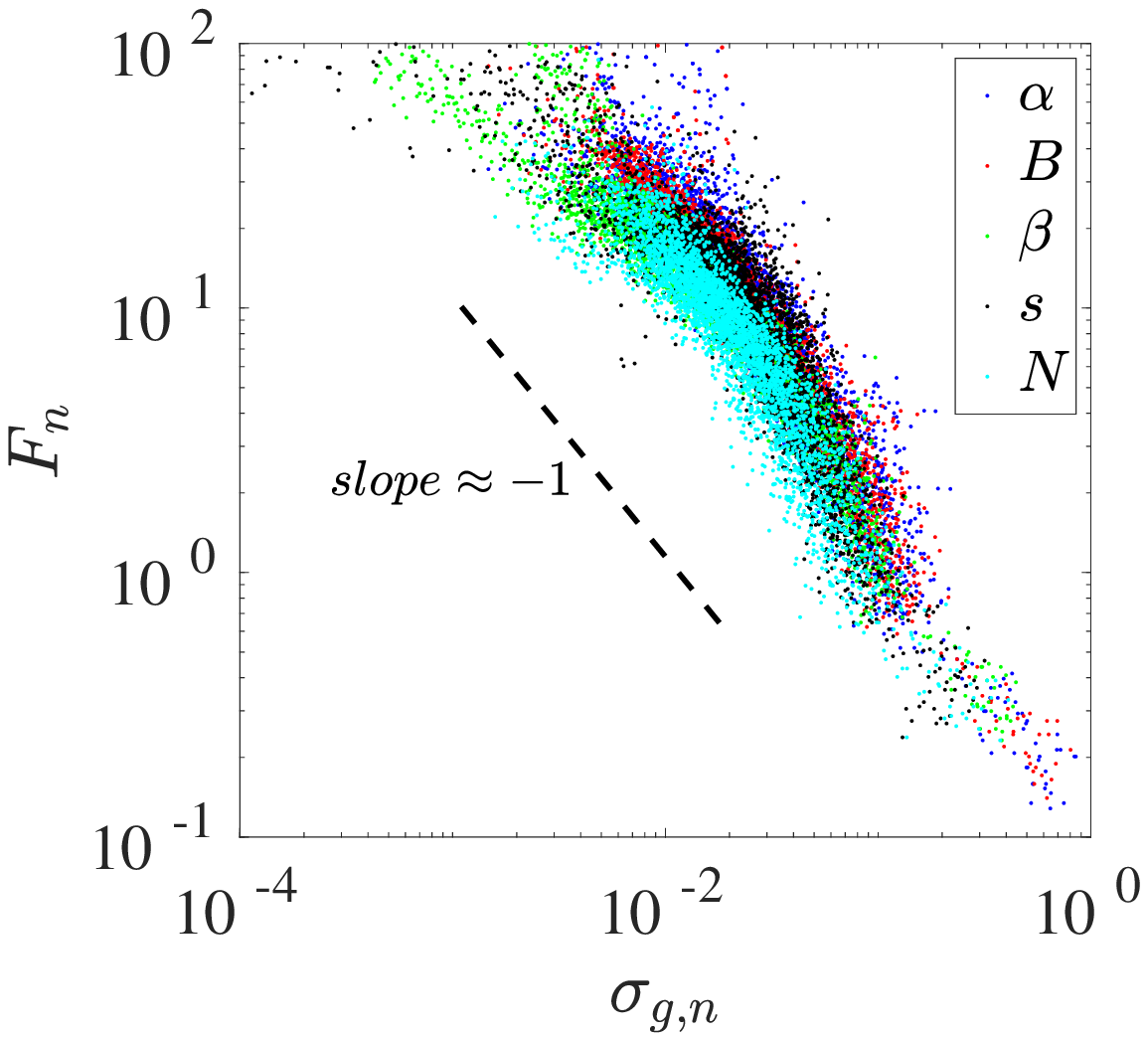}
\caption{Flatness $F_n$ versus $\sigma^2_{g,n}$. Following \cite{feng2020neural}, the flatness $F_n$ in direction-$n$ is defined as the width of the region along direction-$n$, within which the loss function changes within a factor of $2$. }
\label{fig:flatness_sigma_g}
\end{figure}
\FloatBarrier

\begin{figure}[htbp]
\centering
\includegraphics[width=0.9\linewidth]{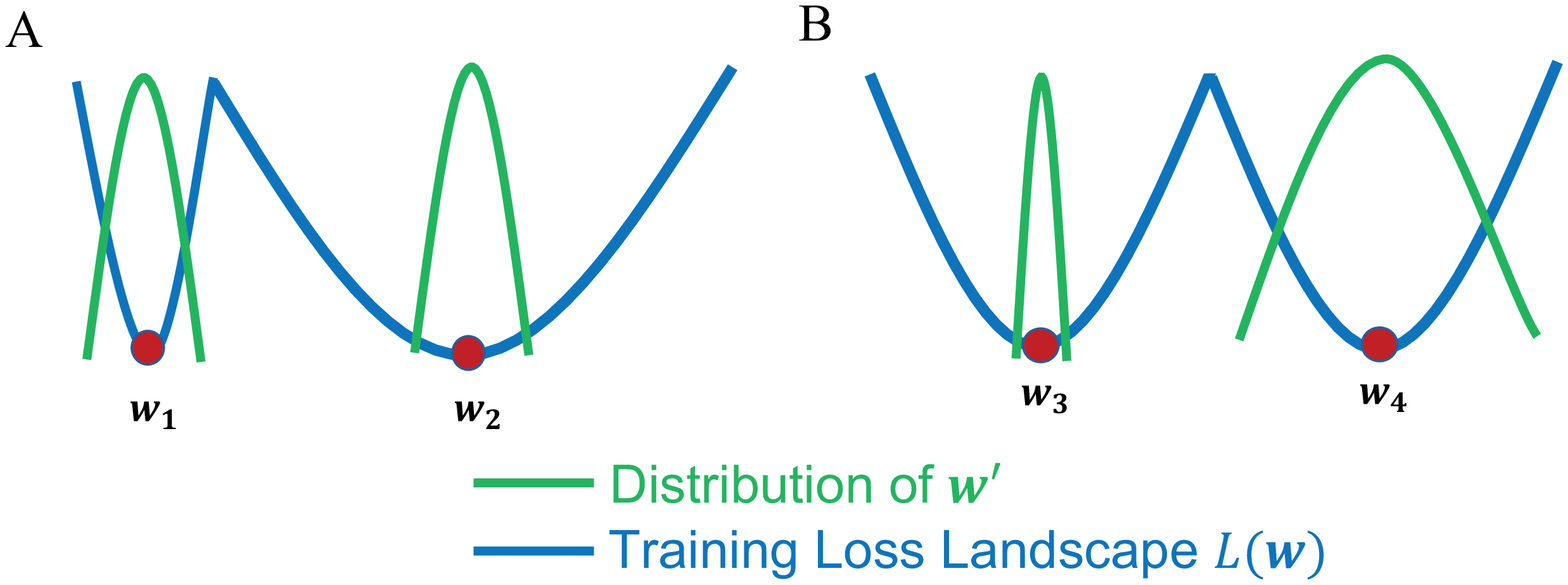}
\caption{Illustration of the two strategies for reducing generalization gap. (A) The flatter solution ($\boldsymbol{w}_2$) has a smaller generalization gap than the sharper solution ($\boldsymbol{w}_1$) if the variances of their dual weights ($\boldsymbol{w}'$) are the same: $\Delta L(\boldsymbol{w}_2)<\Delta L(\boldsymbol{w}_2)$. (B) The smaller solution ($\boldsymbol{w}_3$) with a smaller dual weight variance has a smaller generalization gap than the bigger solution ($\boldsymbol{w}_4$) if their sharpness are the same: $\Delta L(\boldsymbol{w}_2)<\Delta L(\boldsymbol{w}_2)$. The blue line represents the training loss function $L(\boldsymbol{w})$ where the solutions ($\boldsymbol{w}_1$, $\boldsymbol{w}_2$, $\boldsymbol{w}_3$, $\boldsymbol{w}_4$) are at its minima. The green line represents the distribution of the dual weight ($\boldsymbol{w}'$) with its variance $\sigma_{w,n}$ characterizing the size of the solution.   }
\label{fig:lr2} 
\end{figure}

\begin{figure}[htbp]
\centering
\includegraphics[width=0.6\linewidth]{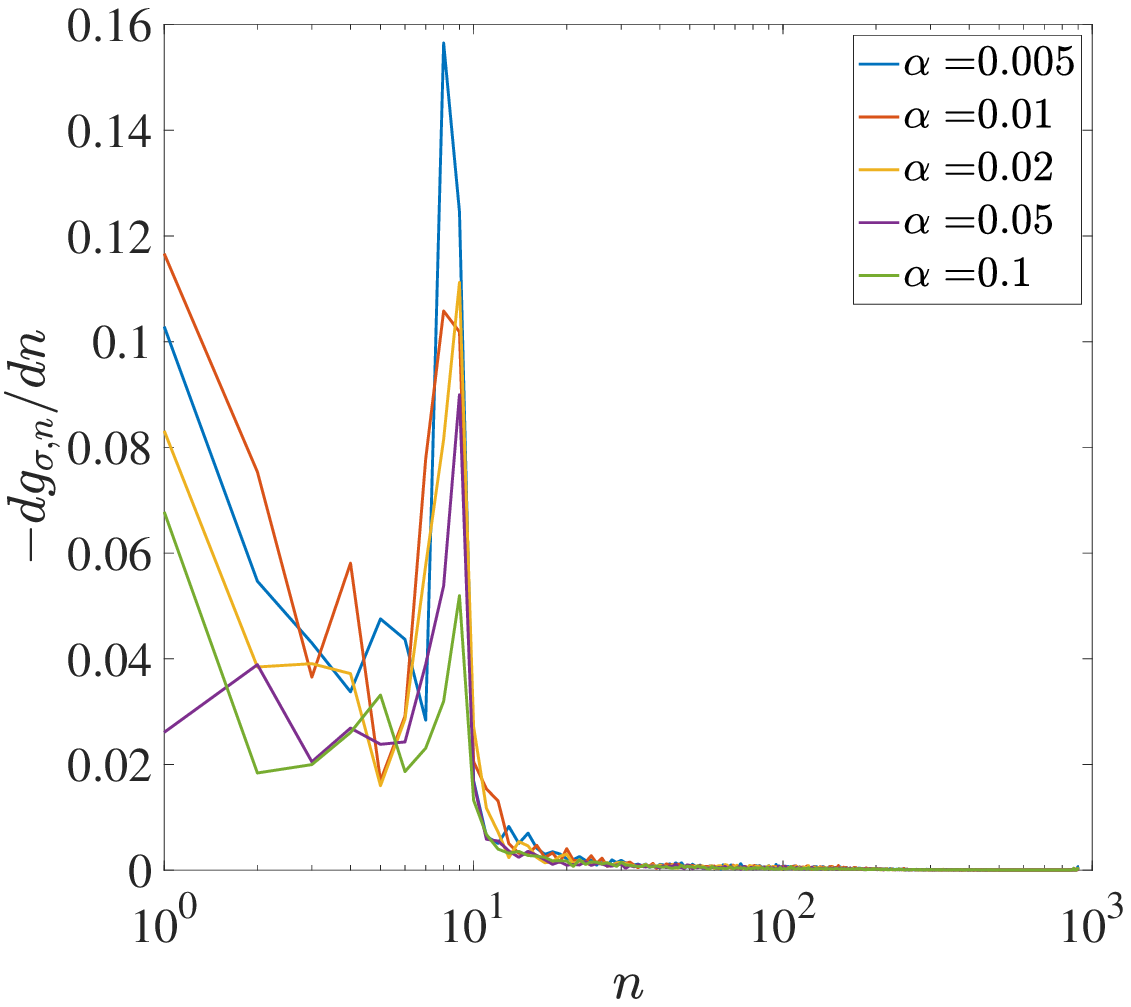}
\caption{ Separation of sharp directions and flat directions. Each point is the average value with ten independent realization.}
\label{fig:sharp_peak}
\end{figure}
\FloatBarrier

\begin{figure}[htbp]
\centering
\includegraphics[width=1\linewidth]{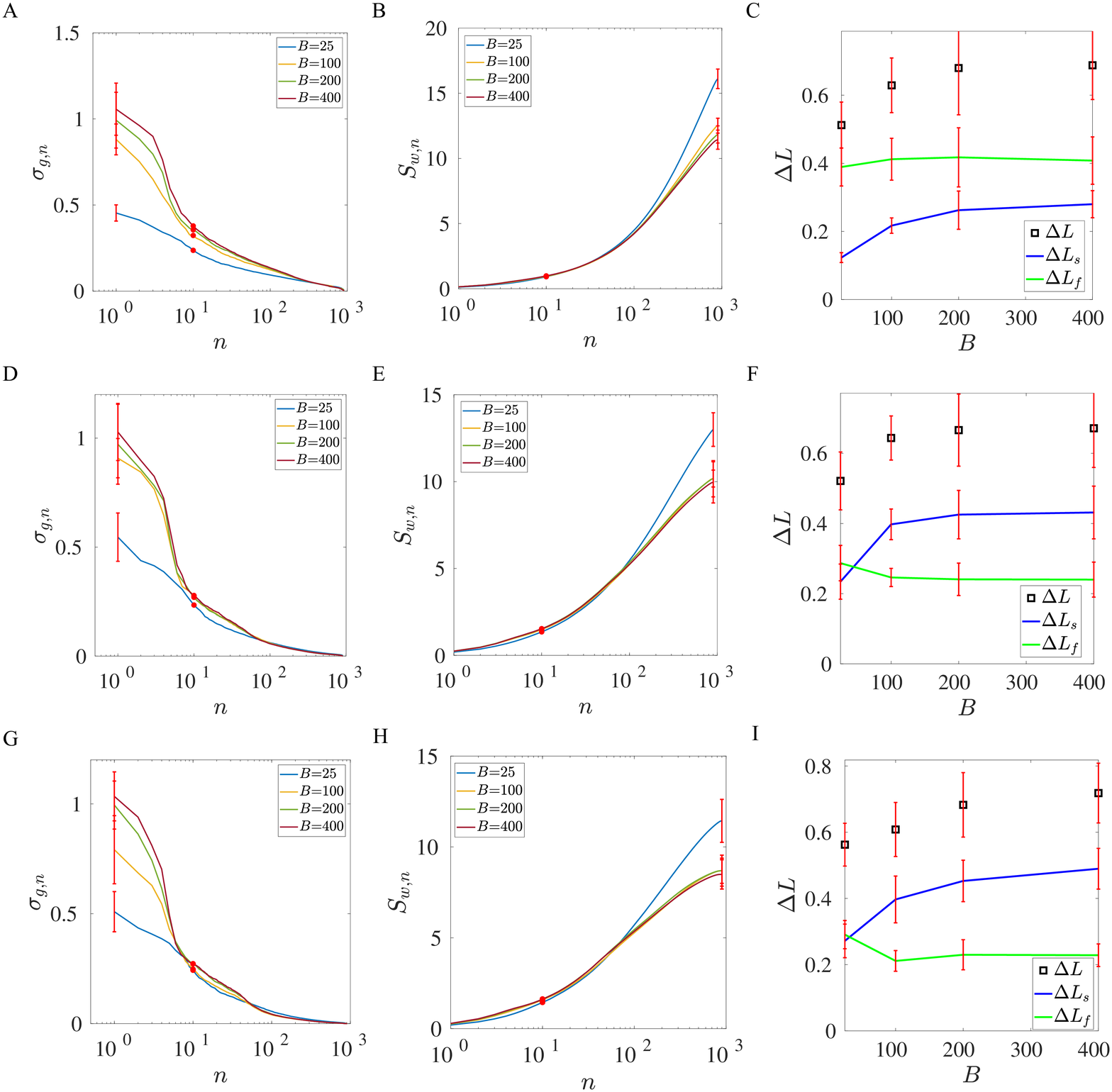}
\caption{The effects of changing batch size $B$ in a multilayer fully connected network. Each row represents a hidden layer: (A-C), (D-F), and (G-I) show the sharpness spectrum $\sigma_{g,n}$, the accumulative weight variation $S_{w,n}=\sum_{i=1}^n \sigma_{w,i}$, and the generalization gap ($\Delta L$) and the contributions from the sharp and flat directions ($\Delta L_s$ and $\Delta L_f$), for hidden layer 1, 2, and 3, respectively.
}
\label{fig:multilayer_B}
\end{figure}

\begin{figure}[htbp]
\centering
\includegraphics[width=1\linewidth]{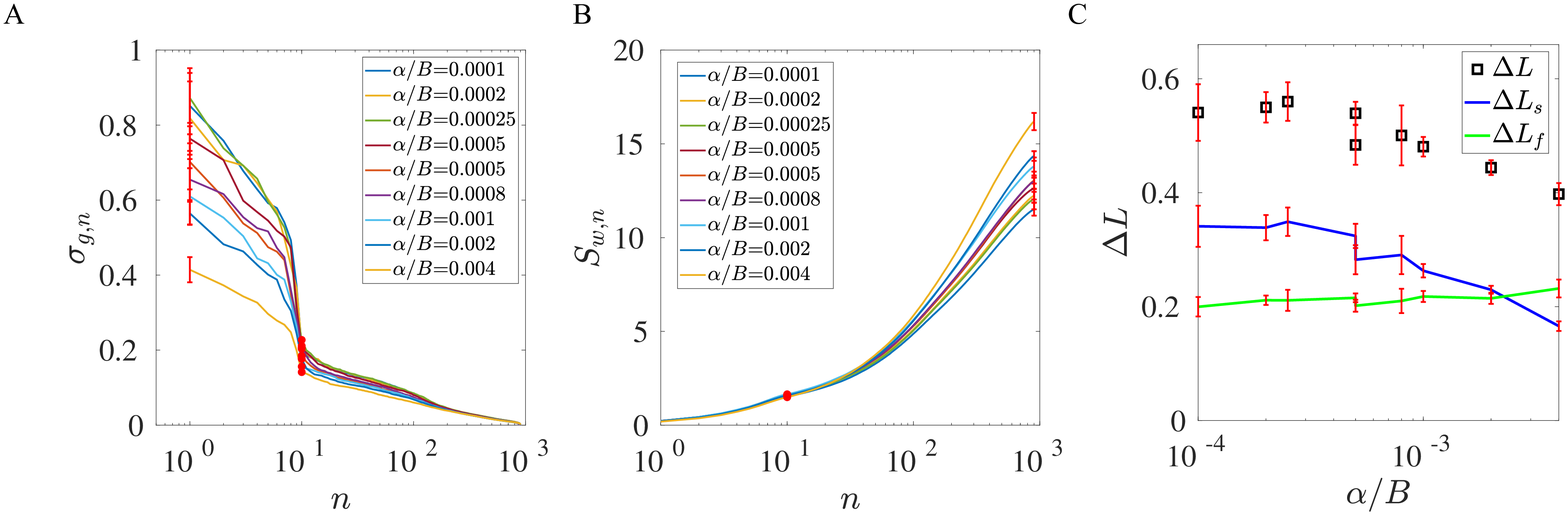}
\caption{The effects of changing both learning rate $\alpha$ and batch size $B$ ($B$ = [25,100,200], $\alpha$ = [0.02,0.05,0.1]).  (A) The sharpness spectrum $\sigma_{g,n}$ for different values of $\alpha/B$. (B) The accumulative size $S_{w,n}=\sum_{i=1}^n \sigma_{w,i}$ for different values of $\alpha/B$. (C) The generalization gap ($\Delta L$) and its contributions from the sharp and flat directions ($\Delta L_s$ and $\Delta L_f$) versus $\alpha/B$. }
\label{fig:flatness_alpha_B}
\end{figure}

\begin{figure}[htbp]
\centering
\includegraphics[width=1\linewidth]{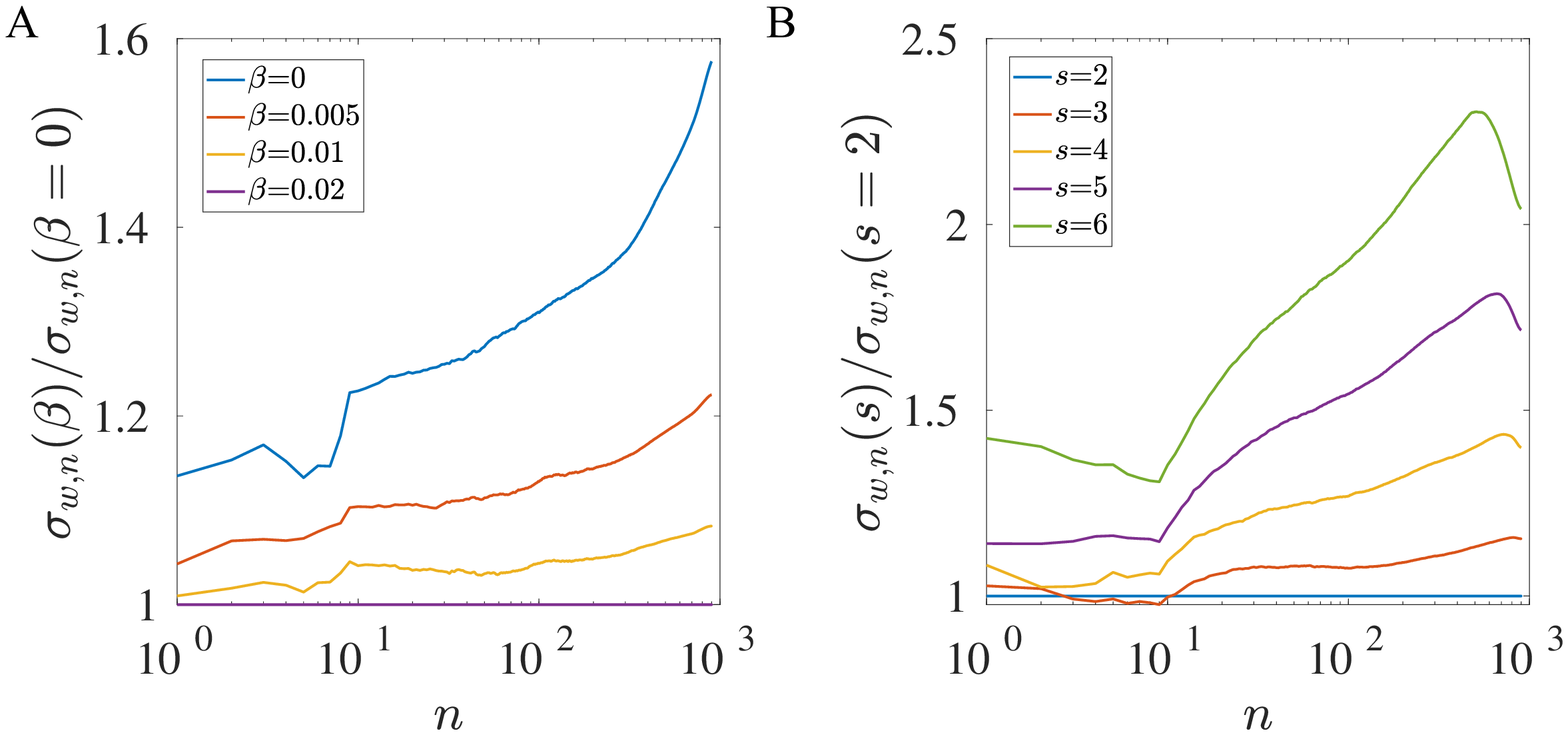}
\caption{The ratio of the weight variance for different weight decay rate and the weight initialization. (A) $\frac{\sigma_{w,n}(\beta)}{\sigma_{w,n}(\beta=0)}$; (B) $\frac{\sigma_{w,n}(s)}{\sigma_{w,n}(s=2)}$. 
}
\label{fig:ratio_sigma_w}
\end{figure}

\begin{figure}[htbp]
\centering
\includegraphics[width=1\linewidth]{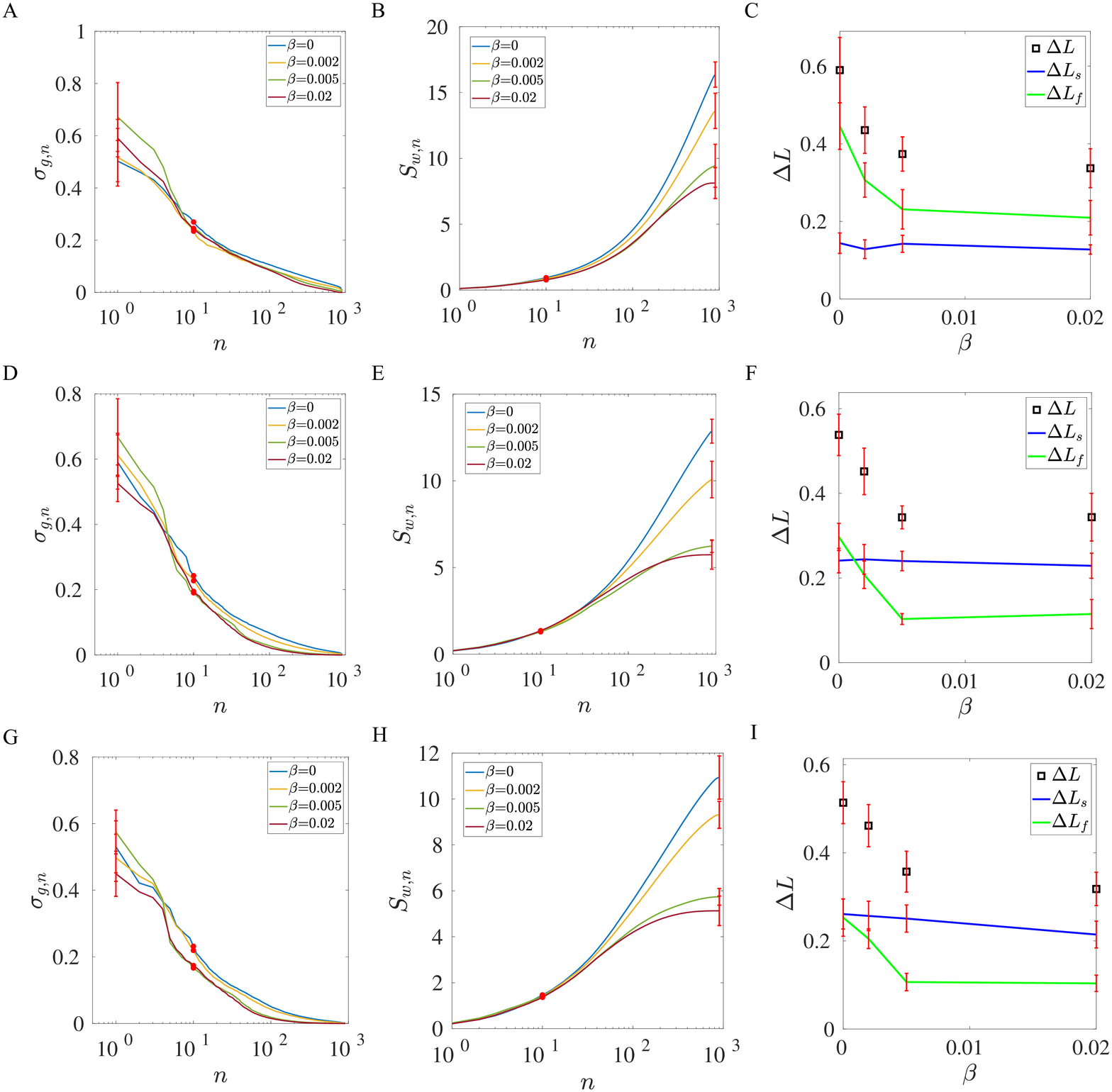}
\caption{The effects of changing weight decay rate  $\beta$ in a multilayer fully connected network. Each row represents a hidden layer: (A-C), (D-F), and (G-I) show the sharpness spectrum $\sigma_{g,n}$, the accumulative weight variation $S_{w,n}=\sum_{i=1}^n \sigma_{w,i}$, and the generalization gap ($\Delta L$) and the contributions from the sharp and flat directions ($\Delta L_s$ and $\Delta L_f$), for hidden layer 1, 2, and 3, respectively. 
}
\label{fig:multilayer_beta}
\end{figure}

\begin{figure}[htbp]
\centering
\includegraphics[width=1\linewidth]{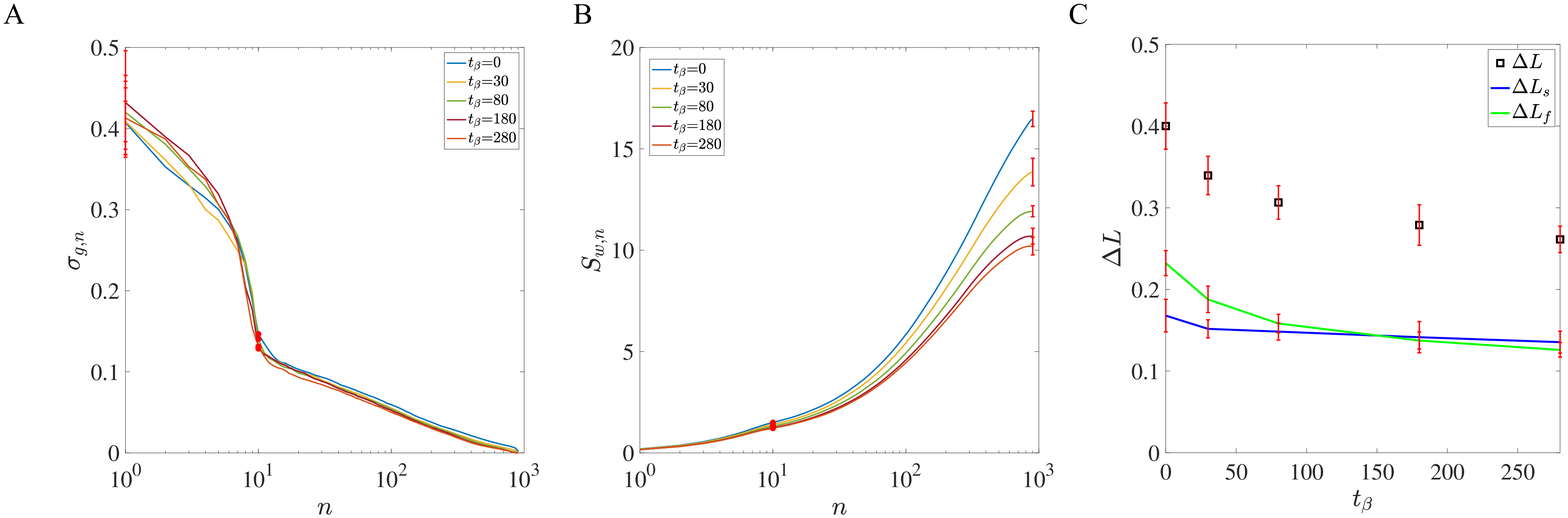}
\caption{ Results with different weight decaying time $t_\beta$ with the MNIST dataset. (A) The sharpness spectrum $\sigma_{g,n}$ does not depend on $t_\beta$ significantly. (B) The accumulative size of the solution $S_{w,n}$ decreases with $t_\beta$ but the change slows down for $t_\beta\ge 100~epoch$. (C) The generalization gap gap $\Delta L$ saturates for $t_\beta\ge 200~epoch$. Weight decay (WD) was applied up to time $t_\beta$ and training is continued without WD until the training loss reaches a preset low threshold ($5\times 10^{-4}$ for MNIST).}
\label{fig:t_beta}
\end{figure}

\begin{figure}[htbp]
\centering
\includegraphics[width=1\linewidth]{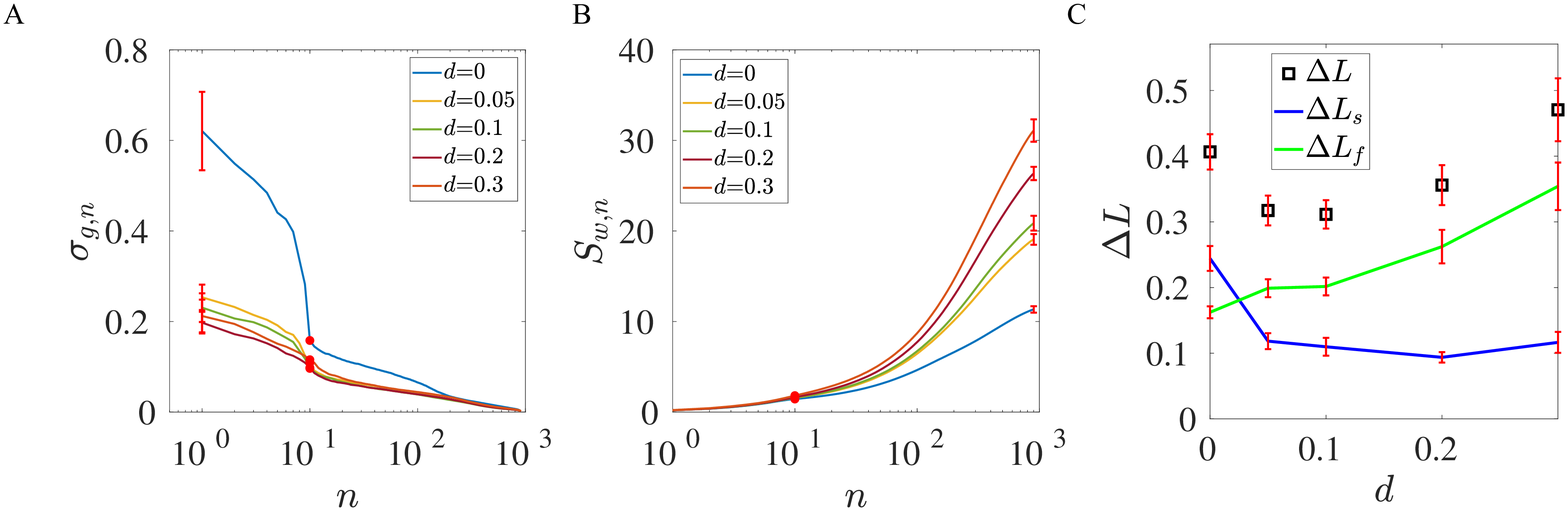}
\caption{ The effects of changing dropout fraction $d$. (A) The sharpness spectrum $\sigma_{g,n}$. (B) The accumulative size $S_{w,n}=\sum_{i=1}^n \sigma_{w,i}$. (C) The generalization gap $\Delta L$ and the contributions from the sharp and flat directions, $\Delta L_s$ and $\Delta L_f$, respectively. All components are ordered by the decreasing order of $\sigma_{g,n}$ (from sharp directions to flat directions). }
\label{fig:dropout}
\end{figure}

\begin{figure}[htbp]
\centering
\includegraphics[width=1.0\linewidth]{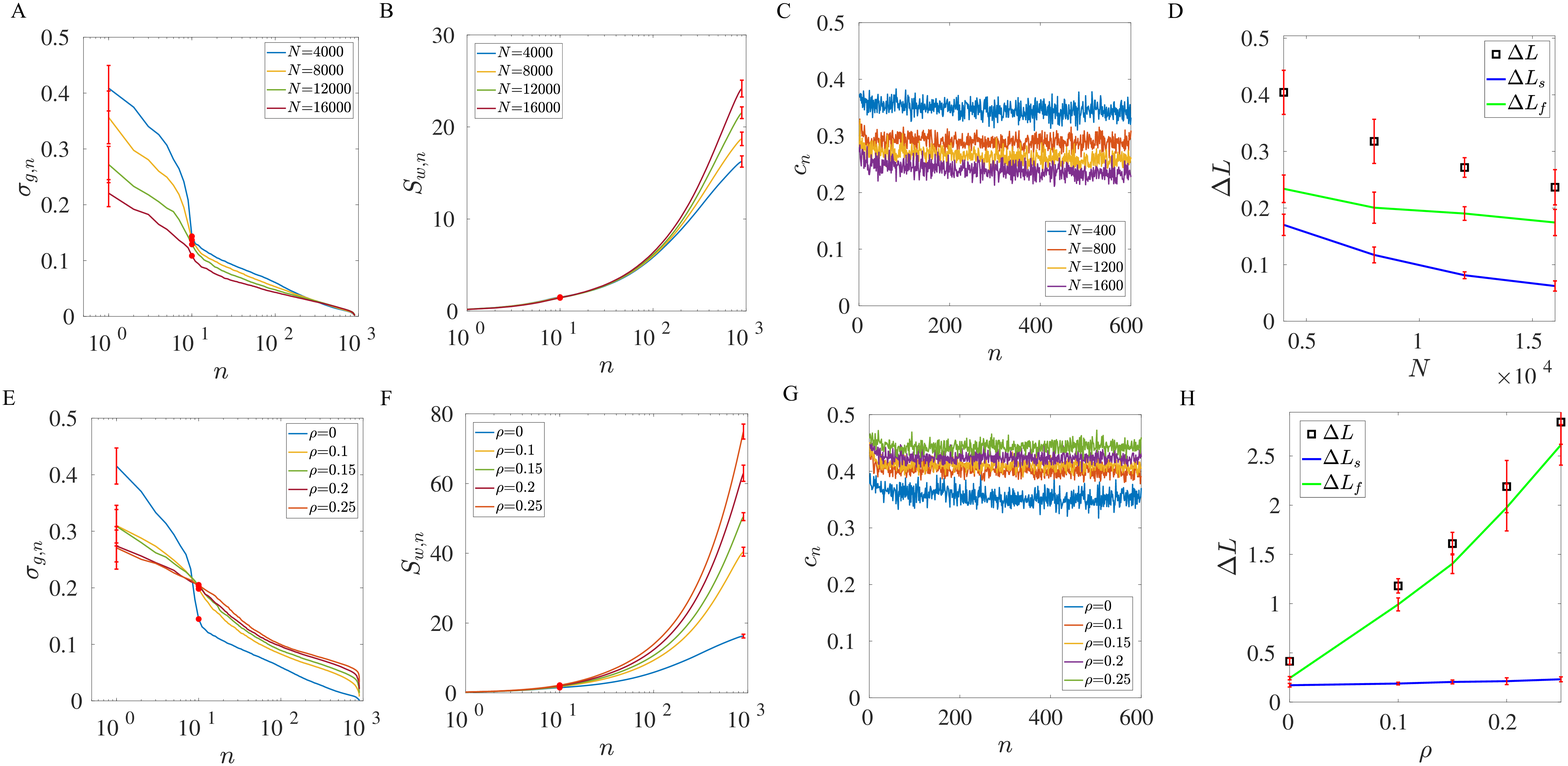}
\caption{The effects of the data set size $N$ (A-D) and the mislabeled fraction $\rho$ (E-H). (A) The sharpness ($\sigma_{g,n}$) decreases in all directions as $N$ increases. (B) The accumulative size $S_{w,n}$ increases slightly. (C) The correlation coefficient $c_n$ decreases with $N$. (D) The decrease in generalization gap ($\Delta L$) comes from both the sharp directions ($\Delta L_s$) and the flat directions ($\Delta L_f)$.  (E) For a finite $\rho=0.1,0.15,0.2,0.25$, the sharpness ($\sigma_{g,n}$) increases significantly in the flat directions ($n>n_s$) while it decreases in the sharp directions, which makes the sharpness spectrum continuous (smooth) without a sudden jump as in the case with $\rho=0$ (blue line). (F) The size of the solution $S_{w,n}$ increases significantly with $\rho$. (G) The correlation coefficient $c_n$ increases with $\rho$.(H) The increase in the overall generalization gap $\Delta L$ with $\rho$ comes dominantly from the flat directions $\Delta L_f$ (green line).}
\label{fig:N_rho}
\end{figure}

\FloatBarrier

\end{document}